\documentclass{article}

\usepackage[preprint, nonatbib]{neurips_2022}

\usepackage[utf8]{inputenc} 
\usepackage[T1]{fontenc}    
\usepackage[pdfborder={0 0 0}]{hyperref}       
\usepackage{url}            
\usepackage{booktabs}       
\usepackage{amsfonts}       
\usepackage{nicefrac}       
\usepackage{microtype}      
\usepackage{xcolor}         

\usepackage{preamble}

\usepackage[sorting=none]{biblatex}
\title{Quaternion Backpropagation}

\author{%
  Johannes Pöppelbaum \\
  Department of Automation Technology\\
  South Westphalia University of Applied Science\\
  59494 Soest, Germany \\
  \texttt{poeppelbaum.johannes@fh-swf.de} \\
   \And
   Andreas Schwung \\
   Department of Automation Technology\\
   South Westphalia University of Applied Science\\
   59494 Soest, Germany \\
   \texttt{schwung.andreas@fh-swf.de} \\
}

\newtheorem{example}{Example}

\begin{document}

\maketitle

\begin{abstract}
%
  Quaternion valued neural networks experienced rising popularity and interest from researchers in the last years, whereby the derivatives with respect to quaternions needed for optimization are calculated as the sum of the partial derivatives with respect to the real and imaginary parts. However, we can show that product- and chain-rule does not hold with this approach. We solve this by employing the GHR-Calculus and derive quaternion backpropagation based on this. Furthermore, we experimentally prove the functionality of the derived quaternion backpropagation.
\end{abstract}

\section{Introduction}

Whist there is a big success and broad range of applications of real-valued neural networks, in the recent years along complex valued models \cite{hirose_complex-valued_2013, trabelsi_deep_2017, benvenuto_complex_1992, ishizuka_modeling_2018, popa_complex-valued_2017, hayakawa_applying_2018} also quaternion valued models \cite{zhu_quaternion_2019, vedaldi_3d-rotation-equivariant_2020, parcollet_quaternion_2016, parcollet_speech_2018, gaudet_deep_2018, onyekpe_quaternion_2021} strike interest of researchers and gain more and more popularity. Applications are e.g. image tasks, 3D point cloud processing, speech / language tasks or sensor fusion. To train these quaternion valued neural networks by means of gradient descent, \cite{nitta_quaternary_1995, parcollet_quaternion_2018, matsui_quaternion_2004} introduced quaternion backpropagation variants, using partial derivatives with respect to the four quaternion components. However, we can show that by doing so, the product rule and chain rule does not hold (compare \ref{subsubsec:simple_partial_derivatives}). This is of particular importance as backpropagation heavily relies on using the chain rule. In this paper, we aim to overcome this issue by proposing a novel quaternion backpropagation, based on the GHR Calculus introduced in \cite{xu_enabling_2015}.

The main contributions of our work are the following:
\begin{itemize}
    \item We show that in initial work on quaternion backpropagation the product rule as well as the chain rule does not hold
    \item We derive quaternion backpropagation using the GHR calculus which defines proper product and chain rules
    \item We employ detailed calculations for each intermediate result, giving proper insights on the underlying quaternion math
    \item We experimentally prove the derived results
\end{itemize}

The paper is structured as follows: Section \ref{sec:related work} presents the related work and Section \ref{sec:fundamentals} the required fundamentals. Section \ref{sec:quaternion_backpropagation} continues with the derived quaternion valued backpropagation algorithm and Section \ref{sec:experiments} provides an experimental proof of it. The paper is concluded by Section \ref{sec:conclusion}.

\renewcommand\nomgroup[1]{%
	\item[\bfseries
	\ifstrequal{#1}{O}{Operators}{%
		\ifstrequal{#1}{N}{Number sets}{%
			\ifstrequal{#1}{R}{Real numbers}{%
				\ifstrequal{#1}{Q}{Quaternions}{%
					\ifstrequal{#1}{0}{Other}{}}}}}%
	]}

\nomenclature[O, 01]{$\times$}{Cross product}
\nomenclature[O, 02]{$\cdot$}{Dot product}
\nomenclature[O, 03]{$\circ$}{Hadamard product}

\nomenclature[N, 01]{\(\mathbb{R}\)}{Real numbers}
\nomenclature[N, 02]{\(\mathbb{C}\)}{Complex numbers}
\nomenclature[N, 03]{\(\mathbb{H}\)}{Quaternions}

\nomenclature[R, 01]{$p, q$}{Scalar}
\nomenclature[R, 02]{$\mathbf{p},\mathbf{q}$}{Vector}
\nomenclature[R, 03]{$\mathbf{P},\mathbf{Q}$}{Matrix}

\nomenclature[Q, 01]{$\quaternion{q}$}{Quaternion}
\nomenclature[Q, 02]{$\quaternionComponents{q}$}{Quaternion components}
\nomenclature[Q, 03]{$\quatCompR{q}$}{Scalar part of quaternion $\quaternion{q}$}
\nomenclature[Q, 04]{$ \mathbf{q} = (\quatCompI{q}, \quatCompJ{q}, \quatCompK{q})$}{Vector part of quaternion $\quaternion{q}$}
\nomenclature[Q, 05]{$\quatConj{q}$}{Quaternion conjugate of $\quaternion{q}$}
\nomenclature[Q, 06]{$\quatInvI{q}$, $\quatInvJ{q}$, $\quatInvK{q}$}{Quaternion Involutions}
\nomenclature[Q, 07]{$\quatConjInvI{q}$, $\quatConjInvJ{q}$, $\quatConjInvK{q}$}{Quaternion conjugate Involutions}

\nomenclature[Q, 08]{$\quatVec{p}$, $\quatVec{q}$}{Quaternion Vector}
\nomenclature[Q, 09]{$\quatVec{P}$, $\quatVec{Q}$}{Quaternion Matrix}


\setlength{\nomlabelwidth}{3.15cm}

\printnomenclature
\section{Related work}
\label{sec:related work}

The related work can mainly be divided into two categories: backpropagation techniques, especially in the complex or hypercomplex domain and applications of quaternion valued neural networks.
For the former, the breakthrough of training neural networks using backpropagation started with the famous paper \cite{rumelhart_learning_1986}. As neural networks emerged from the real numbers to the complex models, equivalent complex variants were introduced in e.g. \cite{leung_complex_1991, nitta_extension_1997, la_corte_newtons_2014}. A detailed description of the usage of the CR / Wirtinger Calculus which comes in handy for complex backpropagation can be found in \cite{kreutz-delgado_complex_2009}. Related are also gradient based optimization techniques in the quaternion domain as described in \cite{mandic_quaternion_2011, xu_optimization_2016, xu_enabling_2015}. However, they differ from our work as they don't target multi-layered, percepton based architectures.

For the latter, \cite{zhu_quaternion_2019} composes fully quaternion valued models by combining quaternion convolution and quaternion fully-connected layers to use them on  classification and denoising tasks of color images. \cite{vedaldi_3d-rotation-equivariant_2020} achieve rotational invariance for point-cloud inputs by utilizing quaternion valued inputs and features, the used weights and biases however are real valued matrices. In the context of Language/Speech understanding \cite{parcollet_quaternion_2016} employs a quaternion MLP approach whereby \cite{parcollet_speech_2018} utilizes quaternion CNN and quaternion RNN to outperform real valued counterparts. \cite{gaudet_deep_2018} proposes novel quaternion weight initialization strategies as well as quaternion batch-normalization to perform image classification and segmentation with quaternion valued models. Finally, \cite{onyekpe_quaternion_2021} applies quaternion RNN, specifically Quaternion Gated Recurrent Units, on sensor fusion for navigation and human activity recognition.
\section{Fundamentals}
\label{sec:fundamentals}

This section introduces the required fundamentals, starting with the math of quaternions, followed by quaternion derivatives and finishes with the regular backpropagation algorithm for real numbered neural networks. 

\subsection{Quaternions}

\subsubsection{Quaternion Algebra}

Quaternions were discovered by Hamilton in 1843 \cite{hamilton_ii_1844} as a method to extend the complex numbers to the three dimensional space. For this, a quaternion consists of three parts, one real and three imaginary

\begin{equation}
    \quaternion{q} = \quaternionComponents{q} = \quatCompR{q} + \mathbf{q}
\end{equation}

where $\quatCompR{q}, \quatCompI{q}, \quatCompJ{q}, \quatCompK{q} \in \mathbb{R}$. Often, $\quatCompR{q}$ is referred to as the real part and $\mathbf{q}$ as the vector part.
The imaginary components $\imagI, \imagJ, \imagK$ have the properties

\begin{equation}
\begin{aligned}
    \imagI^2 = \imagJ^2 = \imagK^2 = \imagI\imagJ\imagK = -1 \\
    \imagI\imagJ = +\imagK,~~ \imagJ\imagK = +\imagI,~~ \imagK\imagI = +\imagJ \\
    \imagJ\imagI = -\imagK,~~ \imagK\imagJ = -\imagI,~~ \imagI\imagK = -\imagJ
\end{aligned}
\end{equation}

Similar to the complex numbers, also quaternions have a conjugate:
\begin{equation}
    \quatConj{q} = \quaternionConjComponents{q} = \quatCompR{q} - \mathbf{q}
\end{equation}

A quaternion $\quaternion{q}$ fulfilling $\left\lVert q \right\rVert = \sqrt{\quaternion{q}\quatConj{q}} = \sqrt{\quatCompR{q}^2 + \quatCompI{q}^2 + \quatCompJ{q}^2 + \quatCompK{q}^2} = 1$ is called a unit quaternion and a quaternion $\quaternion{q}$ with $\quatCompR{q} = 0$ is called a pure quaternion.

The addition of two quaternions $\quaternion{x}, \quaternion{y}$ is defined as
\begin{equation}
    \quaternion{x} + \quaternion{y} = (\quatCompR{x} + \quatCompR{y}) + (\quatCompI{x} + \quatCompI{y})i + (\quatCompJ{x} + \quatCompJ{y})j + (\quatCompK{x} + \quatCompK{y})k
\end{equation}

and multiplication as 
\begin{equation}
\begin{aligned}
    x \; y &= \quatCompR{x}\quatCompR{y} - \mathbf{x} \cdot \mathbf{y} +  x_0\mathbf{y}+y_0\mathbf{x} + \mathbf{x} \times \mathbf{y} \\
	&= ( \quatCompR{x} \quatCompR{y} - \quatCompI{x} \quatCompI{y} - \quatCompJ{x} \quatCompJ{y} - \quatCompK{x} \quatCompK{y} )   \\
	&+ ( \quatCompR{x} \quatCompI{y} + \quatCompI{x} \quatCompR{y} + \quatCompJ{x} \quatCompK{y} - \quatCompK{x} \quatCompJ{y} ) \imagI \\
	&+ ( \quatCompR{x} \quatCompJ{y} - \quatCompI{x} \quatCompK{y} + \quatCompJ{x} \quatCompR{y} + \quatCompK{x} \quatCompI{y} ) \imagJ \\
	&+ ( \quatCompR{x} \quatCompK{y} + \quatCompI{x} \quatCompJ{y} - \quatCompJ{x} \quatCompI{y} + \quatCompK{x} \quatCompR{y} ) \imagK 
\end{aligned}
\end{equation}

We further define the quaternion Hadamard-product of two quaternions $x, y$ as 

\begin{equation}
    \quaternion{x} \circ \quaternion{y} = \quatCompR{x}\quatCompR{y} + \quatCompI{x}\quatCompI{y} \imagI + \quatCompJ{x}\quatCompJ{y} \imagJ + \quatCompK{x} \quatCompK{y} \imagK
\end{equation}

which is needed later on.

\subsubsection{Quaternion Involutions}

Of particular importance for the quaternion derivatives are the quaternion selve inverse mappings or involutions, defined as \cite{ell_quaternion_2007} 

\begin{equation}
    \phi(\eta) = q^\eta = \eta q\eta ^{-1} = \eta q\eta^{-1} = -\eta q\eta
\label{equ:quaternionInvolutionDefinition}
\end{equation}

where $\quaternion{\eta}$ is a pure unit quaternion. Using \eqref{equ:quaternionInvolutionDefinition}, we can create the involutions

\begin{equation}
\begin{aligned}
    \quaternion{q} &= q_0 + q_1i + q_2j + q_3k \\
    \quatInvI{q} &= -i\quaternion{q}i = q_0 + q_1i - q_2j - q_3k \\
    \quatInvJ{q} &= -j\quaternion{q}j = q_0 - q_1i + q_2j - q_3k \\
    \quatInvK{q} &= -k\quaternion{q}k = q_0 - q_1i - q_2j + q_3k .\\
\end{aligned}
\label{equ:quaternionInvolution1}
\end{equation}

The corresponding conjugate involutions are
\begin{equation}
\begin{aligned}
    \quatConjInvI{q} &= q_0 - q_1i + q_2j + q_3k \\
    \quatConjInvJ{q} &= q_0 + q_1i - q_2j + q_3k \\
    \quatConjInvK{q} &= q_0 + q_1i + q_2j - q_3k .\\
\end{aligned}
\end{equation}

Utilizing them, different relations can be created which come in handy in the following quaternion valued derivation calculations as they often help in simplifying the calculations and to avoid elaborate term sorting: \cite{sudbery_quaternionic_1979, xu_enabling_2015} 

\begin{equation}
\begin{aligned}
    \quatCompR{q} &= \frac{1}{4} \left( q + \quatInvI{q} + \quatInvJ{q} + \quatInvK{q} \right) 
    &\quatCompI{q} = -\frac{i}{4} \left( q + \quatInvI{q} - \quatInvJ{q} - \quatInvK{q} \right) \\
    \quatCompJ{q} &= -\frac{j}{4} \left( q - \quatInvI{q} + \quatInvJ{q} - \quatInvK{q} \right) 
    &\quatCompK{q} = -\frac{k}{4} \left( q - \quatInvI{q} - \quatInvJ{q} + \quatInvK{q} \right) \\
\end{aligned}
\label{equ:quaternionInvolution2}
\end{equation}

\begin{equation}
\begin{aligned}
    \quatCompR{q} &= \frac{1}{4} \left( \quatConj{q} + \quatConjInvI{q} + \quatConjInvJ{q} + \quatConjInvK{q} \right) 
    &\quatCompI{q} = \frac{i}{4} \left( \quatConj{q} + \quatConjInvI{q} - \quatConjInvJ{q} - \quatConjInvK{q} \right) \\
    \quatCompJ{q} &= \frac{j}{4} \left( \quatConj{q} - \quatConjInvI{q} + \quatConjInvJ{q} - \quatConjInvK{q} \right) 
    &\quatCompK{q} = \frac{k}{4} \left( \quatConj{q} - \quatConjInvI{q} - \quatConjInvJ{q} + \quatConjInvK{q} \right) \\
\end{aligned}
\label{equ:quaternionInvolution3}
\end{equation}

\begin{equation}
\begin{aligned}
    \quatConj{q}     &= \frac{1}{2} \left( -q + \quatInvI{q} + \quatInvJ{q} + \quatInvK{q} \right) 
    &\quatConjInvI{q} = \frac{1}{2} \left( q - \quatInvI{q} + \quatInvJ{q} + \quatInvK{q} \right) \\
    \quatConjInvJ{q} &= \frac{1}{2} \left( q + \quatInvI{q} - \quatInvJ{q} + \quatInvK{q} \right) 
    &\quatConjInvK{q} = \frac{1}{2} \left( q + \quatInvI{q} + \quatInvJ{q} - \quatInvK{q} \right) \\
\end{aligned}
\label{equ:quaternionInvolution4}
\end{equation}

\begin{equation}
\begin{aligned}
    \quaternion{q} &= \frac{1}{2} \left( -\quatConj{q} + \quatConjInvI{q} + \quatConjInvJ{q} + \quatConjInvK{q} \right) 
    &\quatInvI{q}   = \frac{1}{2} \left( \quatConj{q} - \quatConjInvI{q} + \quatConjInvJ{q} + \quatConjInvK{q} \right) \\
    \quatInvJ{q}   &= \frac{1}{2} \left( \quatConj{q} + \quatConjInvI{q} - \quatConjInvJ{q} + \quatConjInvK{q} \right) 
    &\quatInvK{q}   = \frac{1}{2} \left( \quatConj{q} + \quatConjInvI{q} + \quatConjInvJ{q} - \quatConjInvK{q} \right) \\
\end{aligned}
\label{equ:quaternionInvolution5}
\end{equation}


\subsection{Quaternion Derivatives}

\subsubsection{Simple partial derivatives}
\label{subsubsec:simple_partial_derivatives}

For a quaternion valued function $f(\quaternion{q}), \quaternion{q} = \quaternionComponents{q}$, \cite{nitta_quaternary_1995} and \cite{parcollet_quaternion_2018} calculates the derivatives as follows:

\begin{equation}
    \frac{\partial f}{\partial \quaternion{q}} 
= \left(\frac{\partial f}{\partial \quatCompR{q}} + \frac{\partial f}{\partial \quatCompI{q}}\imagI + \frac{\partial f}{\partial \quatCompJ{q}}\imagJ + \frac{\partial f}{\partial \quatCompK{q}}\imagK \right)
\label{equ:naive_quaternion_derivation}
\end{equation}

However, as we will show now, neither the product rule nor the chain rule hold for this appraoch.

\begin{example}
    When deriving a quaternion valued function $f(\quaternion{q}), \quaternion{q} \in \mathbb{H}$ following \eqref{equ:naive_quaternion_derivation}, the product rule does not apply.
    
    \paragraph{Solution:} Consider $f(\quaternion{q}) = \quaternion{q}\quatConj{q} = \quatCompR{q}^2 + \quatCompI{q}^2 + \quatCompJ{q}^2 + \quatCompK{q}^2$ as the function of choice. Then the direct derivation following \eqref{equ:naive_quaternion_derivation} is 
    \begin{equation}
    \begin{split}
        \frac{\partial f}{\partial \quaternion{q}} 
        &= (\frac{\partial f}{\partial \quatCompR{q}} + \frac{\partial f}{\partial \quatCompI{q}}\imagI + \frac{\partial f}{\partial \quatCompJ{q}}\imagJ + \frac{\partial f}{\partial \quatCompK{q}}\imagK) \\
        &= (2\quatCompR{q} + 2\quatCompI{q}\imagI + 2\quatCompJ{q}\imagJ +2\quatCompK{q}\imagK ) \\
        &= 2\quaternion{q}
    \end{split}
    \end{equation}
    
    Using the product rule, we can calculate the same derivation using 
    \begin{equation}
        \frac{\partial}{\partial \quaternion{q}} \quaternion{q}\quatConj{q} = \quaternion{q} \frac{\partial q^*}{\partial \quaternion{q}} + \frac{\partial \quaternion{q}}{\partial \quaternion{q}}\quatConj{q} .
        \label{equ:product_rule_naive_derivation}
    \end{equation}
    Calculating the partial results
    \begin{equation}
        \begin{aligned}
        \frac{\partial \quatConj{q} }{\partial \quaternion{q}} 
        &= \frac{\partial}{\partial \quaternion{q}} \quatCompR{q} - \quatCompI{q} \imagI - \quatCompJ{q} \imagJ - \quatCompK{q} \imagK\\
        &= 1 - \imagI\imagI - \imagJ\imagJ - \imagK\imagK \\
        &= 1 + 1 + 1 + 1\\
        &= 4
        \end{aligned}
    \end{equation}
    and 
    \begin{equation}
        \begin{aligned}
        \frac{\partial \quaternion{q}}{\partial \quaternion{q}} 
        &= \frac{\partial}{\partial \quaternion{q}} \quatCompR{q} + \quatCompI{q} \imagI + \quatCompJ{q} \imagJ + \quatCompK{q} \imagK\\
        &= 1 + \imagI\imagI + \imagJ\imagJ + \imagK\imagK \\
        &= 1 - 1 - 1 - 1\\
        &= -2
        \end{aligned}
    \end{equation}
    
    and inserting back into \eqref{equ:product_rule_naive_derivation} yields 
    
    \begin{equation}
    \begin{aligned}
        \frac{\partial}{\partial \quaternion{q}} \quaternion{q}\quatConj{q} 
        &= \quaternion{q} \frac{\partial \quatConj{q}}{\partial \quaternion{q}} + \frac{\partial \quaternion{q}}{\partial \quaternion{q}}\quatConj{q} \\
        &= 4\quaternion{q} - 2 \quatConj{q} \neq 2\quaternion{q} .
    \end{aligned}
    \end{equation}
\end{example}

\begin{example}
    When deriving a quaternion valued function $f(\quaternion{z}(\quaternion{x}, \quaternion{y}));~ \quaternion{x}, \quaternion{y}, \quaternion{z} \in \mathbb{H}$ following \eqref{equ:naive_quaternion_derivation}, the chain rule also does not hold.
    
    \paragraph{Solution:} Consider $f(\quaternion{q}) = \quaternion{z} \quatConj{z}; ~\quaternion{z}=\quaternion{x}\quaternion{y}$ as the function of choice.
    
    We first start by calculating the derivative without the chain rule:

    \begin{equation}
    \begin{aligned}
        \frac{\partial f}{\partial \quaternion{x}} &= \frac{\partial}{\partial \quaternion{x}} (\quaternion{x}\quaternion{y})\conj{(\quaternion{x}\quaternion{y})} \\
        &= \frac{\partial}{\partial x} \left(\quatCompR{x} + \quatCompI{x} i + \quatCompJ{x} j + \quatCompK{x} k\right) \left(\quatCompR{y} + \quatCompI{y} i + \quatCompJ{y} j + \quatCompK{y} k\right) \\
        &~\left(\quatCompR{y} - \quatCompI{y} i - \quatCompJ{y} j - \quatCompK{y} k\right) \left(\quatCompR{x} - \quatCompI{x} i - \quatCompJ{x} j - \quatCompK{x} k\right) \\
        &= \frac{\partial}{\partial x} \quatCompR{x}^{2} \quatCompR{y}^{2} + \quatCompR{x}^{2} \quatCompI{y}^{2} + \quatCompR{x}^{2} \quatCompJ{y}^{2} + \quatCompR{x}^{2} \quatCompK{y}^{2} + \quatCompI{x}^{2} \quatCompR{y}^{2} + \quatCompI{x}^{2} \quatCompI{y}^{2} + \quatCompI{x}^{2} \quatCompJ{y}^{2} + \quatCompI{x}^{2} \quatCompK{y}^{2} \\
        &+ \quatCompJ{x}^{2} \quatCompR{y}^{2} + \quatCompJ{x}^{2} \quatCompI{y}^{2} + \quatCompJ{x}^{2} \quatCompJ{y}^{2} + \quatCompJ{x}^{2} \quatCompK{y}^{2} + \quatCompK{x}^{2} \quatCompR{y}^{2} + \quatCompK{x}^{2} \quatCompI{y}^{2} + \quatCompK{x}^{2} \quatCompJ{y}^{2} + \quatCompK{x}^{2} \quatCompK{y}^{2} \\
        &= 2 \quatCompR{x} \quatCompR{y}^{2} + 2 \quatCompR{x} \quatCompI{y}^{2} + 2 \quatCompR{x} \quatCompJ{y}^{2} + 2 \quatCompR{x} \quatCompK{y}^{2} 
        + \left(2 \quatCompI{x} \quatCompR{y}^{2} + 2 \quatCompI{x} \quatCompI{y}^{2} + 2 \quatCompI{x} \quatCompJ{y}^{2} + 2 \quatCompI{x} \quatCompK{y}^{2}\right) \imagI \\
        &+ \left(2 \quatCompJ{x} \quatCompR{y}^{2} + 2 \quatCompJ{x} \quatCompI{y}^{2} + 2 \quatCompJ{x} \quatCompJ{y}^{2} + 2 \quatCompJ{x} \quatCompK{y}^{2}\right) \imagJ 
        + \left(2 \quatCompK{x} \quatCompR{y}^{2} + 2 \quatCompK{x} \quatCompI{y}^{2} + 2 \quatCompK{x} \quatCompJ{y}^{2} + 2 \quatCompK{x} \quatCompK{y}^{2}\right) \imagK
    \end{aligned}
    \end{equation}
    
    Now we will use the chain rule $\frac{\partial f}{\partial \quaternion{x}} = \frac{\partial f}{\partial \quaternion{z}} \frac{\partial \quaternion{z}}{\partial \quaternion{x}}$. We first start with the outer equation:
    
    \begin{equation}
    \begin{aligned}
        \frac{\partial f}{\partial \quaternion{z}} &= \frac{\partial}{\partial \quaternion{z}} \quaternion{z}\quatConj{z} \\
         &= \frac{\partial}{\partial z} \quatCompR{z}^2 + \quatCompI{z}^2 + \quatCompJ{z}^2 + \quatCompK{z}^2 \\
         &= 2(\quatCompR{z} + \quatCompI{z} \imagI + \quatCompJ{z} \imagJ + \quatCompK{z} \imagK) \\
         &= 2\quaternion{z}
    \end{aligned}
    \end{equation}

    \begin{equation}
    \begin{aligned}
        \frac{\partial \quaternion{z}}{\partial \quaternion{x}} &= \frac{\partial}{\partial \quaternion{x}} \quaternion{x}\quaternion{y} \\
        &= \frac{\partial}{\partial \quaternion{x}}\quatCompR{x} \quatCompR{y} + \quatCompR{x} \quatCompI{y}\imagI+ \quatCompR{x} \quatCompJ{y}\imagJ+ \quatCompR{x} \quatCompK{y}\imagK+ \quatCompI{x} \quatCompR{y}\imagI- \quatCompI{x} \quatCompI{y} + \quatCompI{x} \quatCompJ{y}\imagK- \quatCompI{x} \quatCompK{y}\imagJ\\
        &+ \quatCompJ{x} \quatCompR{y}\imagJ- \quatCompJ{x} \quatCompI{y}\imagK- \quatCompJ{x} \quatCompJ{y} + \quatCompJ{x} \quatCompK{y}\imagI+ \quatCompK{x} \quatCompR{y}\imagK+ \quatCompK{x} \quatCompI{y}\imagJ- \quatCompK{x} \quatCompJ{y}\imagI- \quatCompK{x} \quatCompK{y} \\
        &= 2(-\quatCompR{y} + \quatCompI{y}\imagI + \quatCompJ{y}\imagJ + \quatCompK{y})\imagK \\
        &= -2 \quatConj{y}
    \end{aligned}
    \end{equation}
    Combining inner and outer derivative yields 
    \begin{equation}
    \begin{aligned}    
        \frac{\partial f}{\partial \quaternion{x}} &= \frac{\partial f}{\partial \quaternion{z}} \frac{\partial \quaternion{z}}{\partial \quaternion{x}} \\
        &= 2\quaternion{z} (-2\quatConj{y}) \\
        &= -4(\quaternion{x}\quaternion{y}\quatConj{y}) \\
        &= - 4 \quatCompR{x} \quatCompR{y}^{2} - 4 \quatCompR{x} \quatCompI{y}^{2} - 4 \quatCompR{x} \quatCompJ{y}^{2} - 4 \quatCompR{x} \quatCompK{y}^{2} 
        +\left(- 4 \quatCompI{x} \quatCompR{y}^{2} - 4 \quatCompI{x} \quatCompI{y}^{2}- 4 \quatCompI{x} \quatCompJ{y}^{2}- 4 \quatCompI{x} \quatCompK{y}^{2} \right)\imagI \\
        &+\left(- 4 \quatCompJ{x} \quatCompR{y}^{2} - 4 \quatCompJ{x} \quatCompI{y}^{2}- 4 \quatCompJ{x} \quatCompJ{y}^{2} - 4 \quatCompJ{x} \quatCompK{y}^{2} \right)\imagJ
        +\left(- 4 \quatCompK{x} \quatCompR{y}^{2} - 4 \quatCompK{x} \quatCompI{y}^{2} - 4 \quatCompK{x} \quatCompJ{y}^{2} - 4 \quatCompK{x} \quatCompK{y}^{2} \right)\imagK
    \end{aligned}
    \end{equation}

    We can clearly see that the results don't match up and hence chain rule does not hold for this way of calculating quaternion derivatives.
\end{example}

\subsubsection{HR-Calculus}

Similar to the CR-Calculus \cite{kreutz-delgado_complex_2009}, \cite{mandic_quaternion_2011} introduces the HR-Calculus as a method to derive quaternion valued functions.This enables deriving holomorphic quaternionic functions as well as nonholomorphic real functions of quaternion variables. The quaternion derivatives are derived as 
\begin{equation}
\begin{aligned}
    \frac{\partial f}{\partial \quaternion{q}} = 
    \frac{1}{4}(\frac{\partial f}{\partial \quatCompR{q}} - \frac{\partial f}{\partial \quatCompI{q}}\imagI - \frac{\partial f}{\partial \quatCompJ{q}}\imagJ - \frac{\partial f}{\partial \quatCompK{q}}\imagK) \\
    \frac{\partial f}{\partial \quatInvI{q}} = 
    \frac{1}{4}(\frac{\partial f}{\partial \quatCompR{q}} - \frac{\partial f}{\partial \quatCompI{q}}\imagI + \frac{\partial f}{\partial \quatCompJ{q}}\imagJ + \frac{\partial f}{\partial \quatCompK{q}}\imagK) \\
    \frac{\partial f}{\partial \quatInvJ{q}} = 
    \frac{1}{4}(\frac{\partial f}{\partial \quatCompR{q}} + \frac{\partial f}{\partial \quatCompI{q}}\imagI - \frac{\partial f}{\partial \quatCompJ{q}}\imagJ + \frac{\partial f}{\partial \quatCompK{q}}\imagK) \\
    \frac{\partial f}{\partial \quatInvK{q}} = 
    \frac{1}{4}(\frac{\partial f}{\partial \quatCompR{q}} + \frac{\partial f}{\partial \quatCompI{q}}\imagI + \frac{\partial f}{\partial \quatCompJ{q}}\imagJ - \frac{\partial f}{\partial \quatCompK{q}}\imagK) .\\
\end{aligned}    
\label{equ:hr_calculus}
\end{equation}

The corresponding conjugate derivatives are defined as

\begin{equation}
\begin{aligned}
    \frac{\partial f}{\partial \quatConj{q}} = 
    \frac{1}{4}(\frac{\partial f}{\partial \quatCompR{q}} + \frac{\partial f}{\partial \quatCompI{q}}\imagI + \frac{\partial f}{\partial \quatCompJ{q}}\imagJ + \frac{\partial f}{\partial \quatCompK{q}}\imagK) \\
    \frac{\partial f}{\partial \quatConjInvI{q}} = 
    \frac{1}{4}(\frac{\partial f}{\partial \quatCompR{q}} + \frac{\partial f}{\partial \quatCompI{q}}\imagI - \frac{\partial f}{\partial \quatCompJ{q}}\imagJ - \frac{\partial f}{\partial \quatCompK{q}}\imagK) \\
    \frac{\partial f}{\partial \quatConjInvJ{q}} = 
    \frac{1}{4}(\frac{\partial f}{\partial \quatCompR{q}} - \frac{\partial f}{\partial \quatCompI{q}}\imagI + \frac{\partial f}{\partial \quatCompJ{q}}\imagJ - \frac{\partial f}{\partial \quatCompK{q}}\imagK) \\
    \frac{\partial f}{\partial \quatConjInvK{q}} = 
    \frac{1}{4}(\frac{\partial f}{\partial \quatCompR{q}} - \frac{\partial f}{\partial \quatCompI{q}}\imagI - \frac{\partial f}{\partial \quatCompJ{q}}\imagJ + \frac{\partial f}{\partial \quatCompK{q}}\imagK) .\\
\end{aligned}
\label{equ:hr_calculus_conjugate}
\end{equation}

\begin{example}
    When deriving a quaternion valued function $f(q), q \in \mathbb{H}$ using \eqref{equ:hr_calculus} and the known product rule from $\mathbb{R}$, the product rule also does not apply.
    
    \paragraph{Solution:} Again consider $f(\quaternion{q}) = \quaternion{q}\quatConj{q} = \quatCompR{q}^2 + \quatCompI{q}^2 + \quatCompJ{q}^2 + \quatCompK{q}^2$ as the function of choice. Then the direct derivation is 
    \begin{equation}
    \begin{split}
        \frac{\partial f}{\partial \quaternion{q}} 
        &= \frac{1}{4}\left(\frac{\partial f}{\partial \quatCompR{q}} - \frac{\partial f}{\partial \quatCompI{q}}\imagI - \frac{\partial f}{\partial \quatCompJ{q}}\imagJ - \frac{\partial f}{\partial \quatCompK{q}}\imagK \right) \\
        &= \frac{1}{4}\left(2\quatCompR{q} - 2\quatCompI{q}\imagI - 2\quatCompJ{q}\imagJ -2\quatCompK{q}\imagK \right) \\
        &= \frac{1}{2}\quatConj{q}
    \end{split}
    \end{equation}
    
    Using the product rule, we can calculate the same derivation using 
    \begin{equation}
        \frac{\partial}{\partial \quaternion{q}} \quaternion{q}\quatConj{q} = \quaternion{q} \frac{\partial \quatConj{q}}{\partial \quaternion{q}} + \frac{\partial \quaternion{q}}{\partial \quaternion{q}}\quatConj{q}
        \label{equ:product_rule_hr_calculus_derivation}
    \end{equation}
    Calculating the partial results
    \begin{equation}
        \frac{\partial \quatConj{q} }{\partial \quaternion{q}} 
        = \frac{1}{4}(1 +\imagI\imagI + \imagJ\imagJ + \imagK\imagK)
        = -\frac{1}{2}
    \end{equation}
    and 
    \begin{equation}
        \frac{\partial \quaternion{q}}{\partial \quaternion{q}} 
        = \frac{1}{4}(1 - \imagI\imagI - \imagJ\imagJ - \imagK\imagK)
        = 1
    \end{equation}
    
    and inserting back into \eqref{equ:product_rule_hr_calculus_derivation} yields 
    \begin{equation}
        \begin{aligned}
        \frac{\partial}{\partial \quaternion{q}} \quaternion{q}\quatConj{q} 
        &= \quaternion{q} \frac{\partial \quatConj{q}}{\partial \quaternion{q}} + \frac{\partial \quaternion{q}}{\partial \quaternion{q}}\quatConj{q} \\
        &= \frac{-1}{2} \quaternion{q} + 1 \quatConj{q} \neq \frac{1}{2}\quatConj{q}
        \end{aligned}
    \end{equation}
\end{example}

\subsubsection{GHR-Calculus}

Unfortunately, the HR-Calculus lacks the validity of the traditional product rule. Hence, \cite{xu_enabling_2015} extends it to the GHR-Calculus, enabling the definition of a novel quaternion product rule. Generalizing Equations \eqref{equ:hr_calculus} and \eqref{equ:hr_calculus_conjugate}, the derivative and conjugate derivative are defined as


\begin{equation}
\begin{aligned}
    \frac{\partial f}{\partial \quaternion{q}^\quaternion{\mu}} = \frac{1}{4}\left(
    \frac{\partial f}{\partial \quatCompR{q}} - 
    \frac{\partial f}{\partial \quatCompI{q}}\imagI^\quaternion{\mu} - 
    \frac{\partial f}{\partial \quatCompJ{q}}\imagJ^\quaternion{\mu}- 
    \frac{\partial f}{\partial \quatCompK{q}}\imagK^\quaternion{\mu}
    \right)
\end{aligned}
\end{equation}

\begin{equation}
\begin{aligned}
    \frac{\partial f}{\partial \quaternion{q}^{\mu*} } = \frac{1}{4}\left(
    \frac{\partial f}{\partial \quatCompR{q}} + 
    \frac{\partial f}{\partial \quatCompI{q}}\imagI^\quaternion{\mu} + 
    \frac{\partial f}{\partial \quatCompJ{q}}\imagJ^\quaternion{\mu} + 
    \frac{\partial f}{\partial \quatCompK{q}}\imagK^\quaternion{\mu}
    \right)
\end{aligned}
\end{equation}

whereby $\quaternion{\mu} \neq 0, ~ \quaternion{\mu} \in \mathbb{H}$. Consequently, for $\quaternion{\mu} = 1, \imagI, \imagJ, \imagK$ we end up with the derivatives from the HR-Calculus.

\paragraph{The Product Rule} 

Furthermore, they define the quaternion product rule as 
 
\begin{equation}
    \frac{\partial(fg)}{\partial \quaternion{q}^\quaternion{\mu}} 
    = f \frac{\partial(g)}{\partial \quaternion{q}^\quaternion{\mu}} + \frac{\partial(f)}{\partial \quaternion{q}^{\quaternion{g}\quaternion{\mu}}} g,~
    \frac{\partial(fg)}{\partial \quaternion{q}^{\quaternion{\mu} *}} 
    = f \frac{\partial(g)}{\partial \quaternion{q}^{\quaternion{\mu} *}} + \frac{\partial(f)}{\partial \quaternion{q}^{\quaternion{g}\quaternion{\mu} *}} g .
\end{equation}

\paragraph{The Chain Rule}

Finally, \cite{xu_enabling_2015} also defines a quaterniary chain rule as 

\begin{equation}
\begin{aligned}
    \frac{\partial f(g(q))}{\partial q^\mu} &= 
    \frac{\partial f}{\partial g^{\nu}} \frac{\partial g^{\nu}}{\partial q^\mu} + 
    \frac{\partial f}{\partial g^{\nu i}} \frac{\partial g^{\nu i}}{\partial q^\mu} + 
    \frac{\partial f}{\partial g^{\nu j}} \frac{\partial g^{\nu j}}{\partial q^\mu} + 
    \frac{\partial f}{\partial g^{\nu k}} \frac{\partial g^{\nu k}}{\partial q^\mu} \\ 
    \frac{\partial f(g(q))}{\partial q^{\mu*}} &= 
    \frac{\partial f}{\partial g^{\nu}} \frac{\partial g^{\nu}}{\partial q^{\mu*}} + 
    \frac{\partial f}{\partial g^{\nu i}} \frac{\partial g^{\nu i}}{\partial q^{\mu*}} + 
    \frac{\partial f}{\partial g^{\nu j}} \frac{\partial g^{\nu j}}{\partial q^{\mu*}} + 
    \frac{\partial f}{\partial g^{\nu k}} \frac{\partial g^{\nu k}}{\partial q^{\mu*}} 
\end{aligned}
\label{equ:quaternion_chain_rule_1}
\end{equation}

\begin{equation}
\begin{aligned}
    \frac{\partial f(g(q))}{\partial q^\mu} &= 
    \frac{\partial f}{\partial g^{\nu *}} \frac{\partial g^{\nu *}}{\partial q^\mu} + 
    \frac{\partial f}{\partial g^{\nu i*}} \frac{\partial g^{\nu i*}}{\partial q^\mu} + 
    \frac{\partial f}{\partial g^{\nu j*}} \frac{\partial g^{\nu j*}}{\partial q^\mu} + 
    \frac{\partial f}{\partial g^{\nu k*}} \frac{\partial g^{\nu k*}}{\partial q^\mu} \\ 
    \frac{\partial f(g(q))}{\partial q^{\mu*}} &= 
    \frac{\partial f}{\partial g^{\nu *}} \frac{\partial g^{\nu *}}{\partial q^{\mu*}} + 
    \frac{\partial f}{\partial g^{\nu i*}} \frac{\partial g^{\nu i*}}{\partial q^{\mu*}} + 
    \frac{\partial f}{\partial g^{\nu j*}} \frac{\partial g^{\nu j*}}{\partial q^{\mu*}} + 
    \frac{\partial f}{\partial g^{\nu k*}} \frac{\partial g^{\nu k*}}{\partial q^{\mu*}} 
\end{aligned}
\label{equ:quaternion_chain_rule_2}
\end{equation}

with $\quaternion{\mu}, \quaternion{\nu} \in \mathbb{H},~ \quaternion{\mu} \quaternion{\nu}  \neq 0$.

Note that unless otherwise stated, in the following we always use $\quaternion{\mu} = \quaternion{\nu} = 1 + 0\imagI + 0\imagJ + 0\imagK$ as this simplifies the notation throughout the calculations.

\begin{example}
    When using the GHR-Calculus and definitions, product rule can be used as follows: 
    
    \paragraph{Solution:} Again consider $f(\quaternion{q}) = \quaternion{q}\quatConj{q} = \quatCompR{q}^2 + \quatCompI{q}^2 + \quatCompJ{q}^2 + \quatCompK{q}^2$ as the function of choice. Then the direct derivation is 
    \begin{equation}
    \begin{split}
        \frac{\partial f}{\partial \quaternion{q}} 
        &= \frac{1}{4}\left(\frac{\partial f}{\partial \quatCompR{q}} - \frac{\partial f}{\partial \quatCompI{q}}\imagI - \frac{\partial f}{\partial \quatCompJ{q}}\imagJ - \frac{\partial f}{\partial \quatCompK{q}}\imagK \right) \\
        &= \frac{1}{4}\left(2\quatCompR{q} - 2\quatCompI{q}\imagI - 2\quatCompJ{q}\imagJ -2\quatCompK{q}\imagK \right) \\
        &= \frac{1}{2}\quatConj{q}
    \end{split}
    \end{equation}
    
    Using the product rule, we can calculate the same derivation using 
    \begin{equation}
    \frac{\partial(\quaternion{q}\quatConj{q})}{\partial \quaternion{q}^\quaternion{\mu}} 
    = \quaternion{q} \frac{\partial(\quatConj{q})}{\partial \quaternion{q}^\quaternion{\mu}} + \frac{\partial(\quaternion{q})}{\partial \quaternion{q}^{\quatConj{q}\quaternion{\mu}}} \quatConj{q}
    = \quaternion{q} \frac{\partial(\quatConj{q})}{\partial \quaternion{q}} + \frac{\partial(\quaternion{q})}{\partial \quaternion{q}^{\quatConj{q}}} \quatConj{q}
    ~\text{with}~\quaternion{\mu} = 1
    \label{equ:product_rule_ghr_calculus_derivation}
    \end{equation}
    Calculating the partial results
    
    \begin{equation}
    \frac{\partial(\quatConj{q})}{\partial \quaternion{q}^\quaternion{\mu}}
    = \frac{-1}{2}~~ (\text{same as for HR with } \mu = 1)
 \end{equation}
    
    and 
    
    \begin{equation}
    \begin{aligned}
    \frac{\partial(\quaternion{q})}{\partial \quaternion{q}^{\quatConj{q}}} 
     &= \frac{1}{4}\left(\frac{\partial f}{\partial \quatCompR{q}} - \frac{\partial f}{\partial \quatCompI{q}}\imagI^{\quatConj{q}} - \frac{\partial f}{\partial \quatCompJ{q}}\imagJ^{\quatConj{q}} - \frac{\partial f}{\partial \quatCompK{q}}\imagK^{\quatConj{q}} \right) \\
    &= \frac{1}{4} \left( 1 - \imagI\imagI^{\quatConj{q}} - \imagJ\imagJ^{\quatConj{q}} - \imagK\imagK^{\quatConj{q}} \right) \\
    &= \frac{1}{4} \left( \quatConj{q}\quaternion{q}^{*^{-1}} - \imagI \quatConj{q} \imagI \quaternion{q}^{*^{-1}} - \imagJ \quatConj{q} \imagJ \quaternion{q}^{*^{-1}} - \imagK \quatConj{q} \imagK \quaternion{q}^{*^{-1}} \right) \\
    &= \frac{1}{4} \left( \quatConj{q} - \imagI \quatConj{q} \imagI - \imagJ \quatConj{q} \imagJ - \imagK \quatConj{q} \imagK \right) \quaternion{q}^{*^{-1}} \\
    &= \frac{1}{4} \left( \quatConj{q} + \quatConjInvI{q} + \quatConjInvJ{q} + \quatConjInvK{q} \right) \quaternion{q}^{*^{-1}} \\
    &= \quatCompR{q} \quaternion{q}^{*^{-1}}
    \end{aligned}
    \end{equation}
    
    and inserting back into \eqref{equ:product_rule_ghr_calculus_derivation} yields 
    \begin{equation}
    \begin{aligned}
        \frac{\partial(\quaternion{q}\quatConj{q})}{\partial \quaternion{q}^\quaternion{\mu}} 
        &= q \frac{\partial(\quatConj{q})}{\partial \quaternion{q}^\quaternion{\mu}} + \frac{\partial(\quaternion{q})}{\partial \quaternion{q}^{\quaternion{q}\mu}} \quatConj{q} \\
        &= \frac{-1}{2}\quaternion{q} + \quatCompR{q} \quaternion{q}^{*^{-1}} \quatConj{q} \\
        &= \frac{-1}{2}\left(\quatCompR{q} + \quatCompI{q}\imagI + \quatCompJ{q}\imagJ + \quatCompK{q}\imagK\right) + \quatCompR{q} \\
        &= \frac{1}{2}\left(\quatCompR{q} - \quatCompI{q}\imagI - \quatCompJ{q}\imagJ - \quatCompK{q}\imagK\right) \\
        &= \frac{1}{2} \quatConj{q} .
    \end{aligned}
    \end{equation}
    
\end{example}




\subsection{Regular Backpropagation}

\begin{figure}[h]
    \centering
    \includegraphics[width=0.9\textwidth]{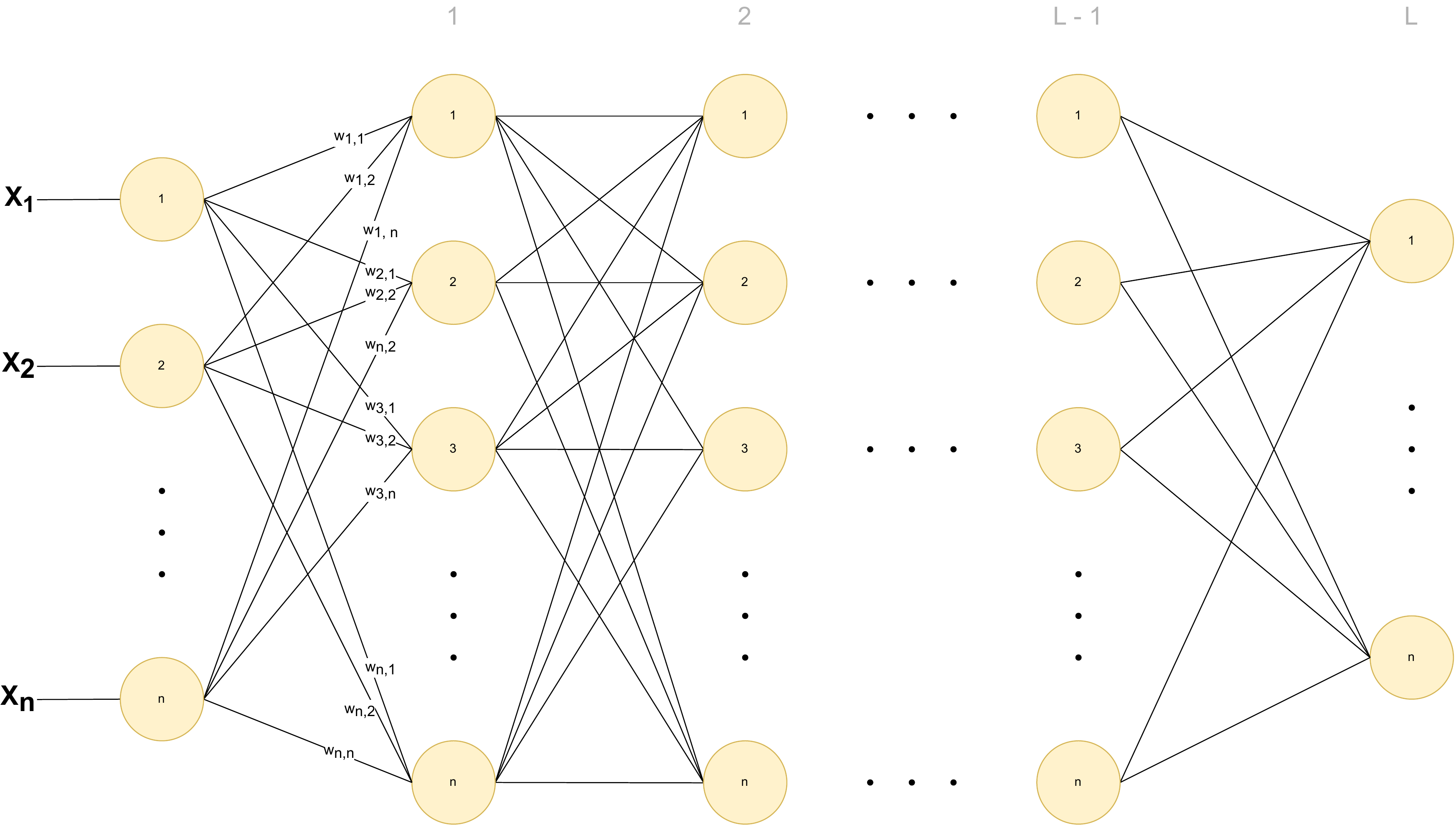}
    \caption{Schematic of a fully connected NN}
    \label{fig:nn_architecture}
\end{figure}

Backpropagation as described in \cite{rumelhart_learning_1986} is usually the go-to method to train NN by minimizing a cost- or loss-function using gradient descent by adjusting its parameters. To understand it, we start with considering the forward phase:
The output of a regular fully connected neural network layer $(l)$ with $n$ inputs and $m$ outputs as depicted in Figure \ref{fig:nn_architecture}
%
can be described with

\begin{equation}
\begin{aligned}
    a^{(l)}_i &= \sigma \left( z^{(l)}_i \right) \\
    z^{(l)}_i &= \sum_{j = 1}^{n} w_{i, j} a^{(l-1)}_j + b^{(l)}_i ~\forall~i \in \left[1, \dots, m\right]
\end{aligned}
\end{equation}

whereby $\mathbf{y} = \mathbf{a}^{(L)}$ is the output of the final layer L and hence the overall network.
This can also conveniently be calculated using the following matrix-vector formulation: 

\begin{equation}
    \begin{bmatrix}
        z^{(l)}_1 \\
        z^{(l)}_2 \\
        \vdots \\
        z^{(l)}_m
    \end{bmatrix} = 
    \begin{bmatrix}
        w^{(l)}_{1,1} & w^{(l)}_{1,2} & \cdots & w^{(l)}_{1,n} \\
        w^{(l)}_{2,1} & w^{(l)}_{2,2} & \cdots & w^{(l)}_{2,n} \\
        \vdots  & \vdots  & \ddots & \vdots    \\
        w^{(l)}_{m,1} & w^{(l)}_{m,2} & \cdots & w^{(l)}_{m,n} \\
    \end{bmatrix}
    \begin{bmatrix}
        a^{(l-1)}_1 \\
        a^{(l-1)}_2 \\
        \vdots \\
        a^{(l-1)}_n
    \end{bmatrix}
\end{equation}

During the training process, the weights $w^{(l)}_{i, j}$ and biases $b^{(l)}_i$ get optimized to minimize the output error function $l(\mathbf{y}, \mathbf{d})$ where $d$ is the desired target output using 

\begin{equation}
\begin{aligned}
    w^{(l)}_{i, j} (n + 1) &=  w^{(l)}_{i, j} (n) - \lambda \frac{\partial \loss}{\partial w^{(l)}_{i, j}} \\
    b^{(l)}_{i} (n + 1) &=  b^{(l)}_{i} (n) - \lambda \frac{\partial \loss}{\partial b^{(l)}_{i}} .
\end{aligned}
\end{equation}

We can formulate the derivatives of a loss function $\loss(\cdot)$ as

\begin{equation}
    \frac{\partial \loss}{\partial w^{(l)}_{i, j}} = \frac{\partial \loss}{\partial a^{(l)}_{i}}
    \frac{\partial a^{(l)}_{i}}{\partial z^{(l)}_{i}} \frac{\partial z^{(l)}_{i}}{\partial w^{(l)}_{i, j}}
\end{equation}
\begin{equation}
    \frac{\partial \loss}{\partial b^{(l)}_{i}} = \frac{\partial \loss}{\partial a^{(l)}_{i}}
    \frac{\partial a^{(l)}_{i}}{\partial z^{(l)}_{i}} \frac{\partial z^{(l)}_{i}}{\partial b^{(l)}_{i}}
\end{equation}

The calculation of $\frac{\partial a^{(l)}_{i}}{\partial z^{(l)}_{i}}$ and $\frac{\partial z^{(l)}_{i}}{\partial w^{(l)}_{i, j}}$ is straight forward, and for $\frac{\partial \loss}{\partial a^{(l)}_{i}}$ we need to differentiate between output and hidden neurons. In case of output neurons, it's likewise straight forward as $\loss(\cdot)$ is a function of $a^{(L)}_j$, for hidden neurons it is 

\begin{equation}
    \frac{\partial \loss}{\partial a^{(l-1)}_{j}} = 
    \sum_{i \in K} \frac{\partial \loss}{\partial a^{(l)}_{i}} \frac{\partial a^{(l)}_{i}}{\partial z^{(l)}_{i}} \frac{\partial z^{(l)}_{i}}{\partial a^{(l-1)}_{j}}
    = \sum_{i \in K} \frac{\partial \loss}{\partial a^{(l)}_{i}} \frac{\partial a^{(l)}_{i}}{\partial z^{(l)}_{i}} w_{i, j} 
\end{equation}

whereby $K = \{1, 2, ... n\}$ indicates all neurons in the following layer $l$ with $\mathbf{a}^{(l-1)}$ as  it's input. Now, starting from the final layer $L$, all derivatives can be calculated in an iterative way backwards to the very first layer.

\section{Quaternion Backpropagation}
\label{sec:quaternion_backpropagation}

In the following, the backpropagation algorithm for quaternion valued neural networks is developed. Initially, the forward phase and loss function is considered, subsequentially this is followed by the derivatives for the last layer of the neural network and finally the derivatives for an arbitrary hidden layer are derived.

\subsection{Forward phase}

We can formulate the forward phase of a regular FFQN layer $(l)$ with $n$ inputs and $m$ outputs as follows:

\begin{equation}
\begin{aligned}
    \quaternion{a}_i^{(l)} &= \sigma(\quaternion{z}_i^{(l)}) \\
    \quaternion{z}_i^{(l)} &= \sum_{j=1}^{n} \quaternion{w}_{i,j}^{(l)} \quaternion{a}_j^{(l-1)} + \quaternion{b}_i^{(l)}
\end{aligned}
\end{equation}

where $i \in \{1, \dots, m\}$, $j \in \{1, \dots, n\}$ and $w, y, b \in \mathbb{Q}$. 
The corresponding matrix-vector formulation is

\begin{equation}
\begin{aligned}
    \quatVec{a}^{(l)} &= \sigma(\quatVec{z}^{(l)}) \\
    \quatVec{z}^{(l)} &= \quatVec{W}^{(l)} \quatVec{a}^{(l-1)} + \quatVec{b}^{(l)}
\end{aligned}
\end{equation}

where $\quatVec{W} \in \mathbb{Q}^{m \times n}$ and $\quatVec{b} \in \mathbb{Q}^{m}$ .The final output $\quatVec{y}$ of the last layer $L$ and hence the overall model is $\quatVec{a}^{(L)}$.

\subsection{Loss function}

We define the loss function $\loss()$ between the desired output $\quaternion{d}_i$ and the actual output $\quaternion{y}_i$ as
\begin{equation}
    \loss = \quatVec{e}^{*^T} \quatVec{e} 
\end{equation}
whereby $\quaternion{e}_i \in \{1, ..., m \}$ and $\quaternion{e}_i = \quaternion{d}_i - \quaternion{y}_i$. By doing so, we obtain the real valued loss 
\begin{equation}
\begin{aligned}
    \loss &= \quatConj{e}_1 \quaternion{e}_1 + \quatConj{e}_2 \quaternion{e}_2 + \dots + \quatConj{e}_m \quaternion{e}_m  \\
    &= ({\quatCompR{e}}_1^2 + {\quatCompI{e}}_1^2 + {\quatCompJ{e}}_1^2 + {\quatCompK{e}}_1^2)
    + ({\quatCompR{e}}_2^2 + {\quatCompI{e}}_2^2 + {\quatCompJ{e}}_2^2 + {\quatCompK{e}}_2^2)
    + \dots
    + ({\quatCompR{e}}_m^2 + {\quatCompI{e}}_m^2 + {\quatCompJ{e}}_m^2 + {\quatCompK{e}}_m^2) \\
    &= \left(({\quatCompR{d}}_1 - {\quatCompR{y}}_1)^2 + ({\quatCompI{d}}_1 - {\quatCompI{y}}_1)^2 + ({\quatCompJ{d}}_1 - {\quatCompJ{y}}_1)^2 + ({\quatCompK{d}}_1 - {\quatCompK{y}}_1)^2 \right) \\
    &+ \left(({\quatCompR{d}}_2 - {\quatCompR{y}}_2)^2 + ({\quatCompI{d}}_2 - {\quatCompI{y}}_2)^2 + ({\quatCompJ{d}}_2 - {\quatCompJ{y}}_2)^2 + ({\quatCompK{d}}_2 - {\quatCompK{y}}_2)^2 \right) \\
    &+ \dots \\
    &+ \left(({\quatCompR{d}}_m - {\quatCompR{y}}_m)^2 + ({\quatCompI{d}}_m - {\quatCompI{y}}_m)^2 + ({\quatCompJ{d}}_m - {\quatCompJ{y}}_m)^2 + ({\quatCompK{d}}_m - {\quatCompK{y}}_m)^2 \right)
\end{aligned}
\end{equation}

\subsection{Final layer}
\label{subsec:final_layer}

We start deriving the quaternion backpropagation algorithm by first considering the output of the final layer of a QNN without the usage of an activation function. If an activation function shall be used, one can use intermediate results and apply the strategy from the following Subsection \ref{subsec:hidden_layer}. Consequently, the output $\quatVec{y}$ is calculated as

\begin{equation}
    \quatVec{y} = \quatVec{W}^{(L)} \quatVec{a}^{(L-1)} + \quatVec{b}^{(L)}
\end{equation}
where $\quatVec{a}^{(L-1)}$ is the output of the previous layer. One output quaternion $\quaternion{y}_i$ can be obtained using
\begin{equation}
    \quaternion{y}_i = \sum_{j=1}^{n} \quaternion{w}_{i,j}^{(L)} \quaternion{a}_j^{(L-1)} + \quaternion{b}_i^{(L)} .
\end{equation}

For deriving $\loss$ with respect to the weights $\quatVec{W}^{(L)}$, biases $\quatVec{b}^{(L)}$ and activation outputs of the previous layer $\quatVec{a}^{(L-1)}$, we utilize the chain rule where we first derive $\frac{\partial \loss}{\partial \quaternion{y}}$ and then $\frac{\partial \quaternion{y}}{\partial \conj{\quatVec{W}}}$, $\frac{\partial \quaternion{y}}{\partial \conj{\quatVec{b}}}$ and $\frac{\partial \quaternion{y}}{\partial \quatVec{a}}$.

Note that we use the conjugate $\conj{\quatVec{W}}$ and $\conj{\quatVec{b}}$ as this is the direction of the steepest descent \cite{mandic_quaternion_2011, xu_enabling_2015}. For better readability, we waive on the subscripts $\square_{i,j}$ indicating the matrix/vector elements as well as the superscript $\square^{(L)}$ throughout the following calculations.

\subsubsection{Derivative with respect to weights}

First, we consider the derivative with respect to the weights $\quaternion{w}^{(L)}_{i, j}$

\begin{equation}
    \frac{\partial \loss(\quaternion{y}(\quaternion{w}, \quaternion{b})}{\partial \quatConj{w}} = 
    \quaternionDerivative{\loss}{\quaternion{y}}{\quatConj{w}}.
\end{equation}

We start by calculating the respective left partial derivatives

\begin{equation}
\begin{aligned}
    \frac{\partial \loss}{\partial \quaternion{y}} 
    &= \frac{\partial }{\partial e} (\quatCompR{d} - \quatCompR{y})^2 + (\quatCompI{d} - \quatCompI{y})^2 + (\quatCompJ{d} - \quatCompJ{y})^2 + (\quatCompK{d} - \quatCompK{y})^2 \\ 
    &= \frac{1}{4}\left[-2(\quatCompR{d} - \quatCompR{y}) + 2(\quatCompI{d} - \quatCompI{y})\imagI + 2(\quatCompJ{d} - \quatCompJ{y})\imagJ + 2(\quatCompK{d} - \quatCompK{y})\imagK \right] \\
    &= -\frac{1}{2}\left[(\quatCompR{d} - \quatCompR{y}) - (\quatCompI{d} - \quatCompI{y})\imagI - (\quatCompJ{d} - \quatCompJ{y})\imagJ - (\quatCompK{d} - \quatCompK{y})\imagK \right] \\
    &= -\frac{1}{2} (d - y)^* = -\frac{1}{2} \quatConj{e}
\end{aligned}
\end{equation}

\begin{equation}
\begin{aligned}
    \frac{\partial \loss}{\partial \quatInvI{y}} 
    &= \frac{\partial }{\partial e} (\quatCompR{d} - \quatCompR{y})^2 + (\quatCompI{d} - \quatCompI{y})^2 + (\quatCompJ{d} - \quatCompJ{y})^2 + (\quatCompK{d} - \quatCompK{y})^2 \\ 
    &= \frac{1}{4}\left[-2(\quatCompR{d} - \quatCompR{y}) + 2(\quatCompI{d} - \quatCompI{y})\imagI - 2(\quatCompJ{d} - \quatCompJ{y})\imagJ - 2(\quatCompK{d} - \quatCompK{y})\imagK \right] \\
    &= -\frac{1}{2}\left[(\quatCompR{d} - \quatCompR{y}) - (\quatCompI{d} - \quatCompI{y})\imagI + (\quatCompJ{d} - \quatCompJ{y})\imagJ + (\quatCompK{d} - \quatCompK{y})\imagK \right] \\
    &= -\frac{1}{2} \quatConjInvI{(d - y)} = -\frac{1}{2} \quatConjInvI{e}
\end{aligned}
\end{equation}

\begin{equation}
\begin{aligned}
    \frac{\partial \loss}{\partial \quatInvJ{y}} 
    &= \frac{\partial }{\partial e} (\quatCompR{d} - \quatCompR{y})^2 + (\quatCompI{d} - \quatCompI{y})^2 + (\quatCompJ{d} - \quatCompJ{y})^2 + (\quatCompK{d} - \quatCompK{y})^2 \\ 
    &= \frac{1}{4}\left[-2(\quatCompR{d} - \quatCompR{y}) - 2(\quatCompI{d} - \quatCompI{y})\imagI + 2(\quatCompJ{d} - \quatCompJ{y})\imagJ - 2(\quatCompK{d} - \quatCompK{y})\imagK \right] \\
    &= -\frac{1}{2}\left[(\quatCompR{d} - \quatCompR{y}) + (\quatCompI{d} - \quatCompI{y})\imagI - (\quatCompJ{d} - \quatCompJ{y})\imagJ + (\quatCompK{d} - \quatCompK{y})\imagK \right] \\
    &= -\frac{1}{2} \quatConjInvJ{(d - y)} = -\frac{1}{2} \quatConjInvJ{e}
\end{aligned}
\end{equation}

\begin{equation}
\begin{aligned}
    \frac{\partial \loss}{\partial \quatInvK{y}} 
    &= \frac{\partial }{\partial e} (\quatCompR{d} - \quatCompR{y})^2 + (\quatCompI{d} - \quatCompI{y})^2 + (\quatCompJ{d} - \quatCompJ{y})^2 + (\quatCompK{d} - \quatCompK{y})^2 \\ 
    &= \frac{1}{4}\left[-2(\quatCompR{d} - \quatCompR{y}) - 2(\quatCompI{d} - \quatCompI{y})\imagI - 2(\quatCompJ{d} - \quatCompJ{y})\imagJ + 2(\quatCompK{d} - \quatCompK{y})\imagK \right] \\
    &= -\frac{1}{2}\left[(\quatCompR{d} - \quatCompR{y}) + (\quatCompI{d} - \quatCompI{y})\imagI + (\quatCompJ{d} - \quatCompJ{y})\imagJ - (\quatCompK{d} - \quatCompK{y})\imagK \right] \\
    &= -\frac{1}{2} \quatConjInvK{(d - y)} = -\frac{1}{2} \quatConjInvK{e}
\end{aligned}
\end{equation}

Now we calculate the right partial derivatives of $\quaternion{y}, \quatInvI{y}, \quatInvJ{y}~ \text{and}~\quatInvK{y}$ with respect to $\quatConj{w}$\\

\begin{equation}
\begin{aligned}
    \frac{\partial \quaternion{y}}{\partial \quatConj{w}} 
    &= \frac{\partial (\quaternion{w}\quaternion{a} + \quaternion{b})}{\partial \quatConj{w}}
    = \frac{\partial (\quaternion{w}\quaternion{a})}{\partial \quatConj{w}} \\
    &= \frac{\partial}{\partial \quatConj{w}} 
    (\quatCompR{a} \quatCompR{w} - \quatCompI{a} \quatCompI{w} - \quatCompJ{a} \quatCompJ{w} - \quatCompK{a} \quatCompK{w})
    + (\quatCompR{a} \quatCompI{w} + \quatCompI{a} \quatCompR{w} - \quatCompJ{a} \quatCompK{w} + \quatCompK{a} \quatCompJ{w} ) \imagI \\
    &+ (\quatCompR{a} \quatCompJ{w} + \quatCompI{a} \quatCompK{w} + \quatCompJ{a} \quatCompR{w} - \quatCompK{a} \quatCompI{w} ) \imagJ
    + (\quatCompR{a} \quatCompK{w} - \quatCompI{a} \quatCompJ{w} + \quatCompJ{a} \quatCompI{w} + \quatCompK{a} \quatCompR{w} ) \imagK \\
    &= \frac{1}{4} [ \quatCompR{a} + \quatCompI{a} \imagI + \quatCompJ{a} \imagJ + \quatCompK{a} k + (\quatCompR{a} \imagI - \quatCompI{a} + \quatCompJ{a} \imagK - \quatCompK{a} \imagJ) \imagI \\
    &+ (\quatCompR{a} \imagJ - \quatCompI{a} \imagK - \quatCompJ{a} + \quatCompK{a} i) \imagJ + (\quatCompR{a} \imagK + \quatCompI{a} \imagJ - \quatCompJ{a} \imagI - \quatCompK{a}) \imagK ] \\
    &= \frac{1}{2} \left[ -\quatCompR{a} + \quatCompI{a} + \quatCompJ{a} + \quatCompK{a} \right] \\
    &= - \frac{1}{2} \quatConj{a}
\end{aligned}    
\end{equation}
%
%
\begin{equation}
\begin{aligned}
    \frac{\partial \quatInvI{y}}{\partial \quatConj{w}} 
    &= \frac{\partial \InvI{(\quaternion{w}\quaternion{a} + \quaternion{b})}}{\partial \quatConj{w}}
    = \frac{\partial \InvI{(\quaternion{w}\quaternion{a})}}{\partial \quatConj{w}} \\
    &= \frac{\partial}{\partial \quatConj{w}} 
    (\quatCompR{a} \quatCompR{w} - \quatCompI{a} \quatCompI{w} - \quatCompJ{a} \quatCompJ{w} - \quatCompK{a} \quatCompK{w})
    + (\quatCompR{a} \quatCompI{w} + \quatCompI{a} \quatCompR{w} - \quatCompJ{a} \quatCompK{w} + \quatCompK{a} \quatCompJ{w} ) \imagI \\
    &- (\quatCompR{a} \quatCompJ{w} + \quatCompI{a} \quatCompK{w} + \quatCompJ{a} \quatCompR{w} - \quatCompK{a} \quatCompI{w} ) \imagJ
    - (\quatCompR{a} \quatCompK{w} - \quatCompI{a} \quatCompJ{w} + \quatCompJ{a} \quatCompI{w} + \quatCompK{a} \quatCompR{w} ) \imagK \\
    &= \frac{1}{4} 
    [\quatCompR{a} + \quatCompI{a} \imagI - \quatCompJ{a} \imagJ - \quatCompK{a} \imagK
    + (\quatCompR{a}\imagI - \quatCompI{a} - \quatCompJ{a}\imagK + \quatCompK{a}\imagJ) \imagI \\
    &+ (-\quatCompR{a}\imagJ + \quatCompI{a}\imagK - \quatCompJ{a} + \quatCompK{a}\imagI) \imagJ
    + (-\quatCompR{a}\imagK - \quatCompI{a}\imagJ - \quatCompJ{a}\imagI - \quatCompK{a}) \imagK ] \\
    &= \frac{1}{2} \left[\quaternionConjComponents{a} \right] \\
    &= \frac{1}{2} \quatConj{a}
\end{aligned}    
\end{equation}
%
%
\begin{equation}
\begin{aligned}
    \frac{\partial \quatInvJ{y}}{\partial \quatConj{w}} 
    &= \frac{\partial \InvJ{(\quaternion{w}\quaternion{a} + \quaternion{b})}}{\partial \quatConj{w}}
    = \frac{\partial \InvJ{(\quaternion{w}\quaternion{a})}}{\partial \quatConj{w}} \\
    &= \frac{\partial}{\partial \quatConj{w}} 
    (\quatCompR{a} \quatCompR{w} - \quatCompI{a} \quatCompI{w} - \quatCompJ{a} \quatCompJ{w} - \quatCompK{a} \quatCompK{w})
    - (\quatCompR{a} \quatCompI{w} + \quatCompI{a} \quatCompR{w} - \quatCompJ{a} \quatCompK{w} + \quatCompK{a} \quatCompJ{w} ) \imagI \\
    &+ (\quatCompR{a} \quatCompJ{w} + \quatCompI{a} \quatCompK{w} + \quatCompJ{a} \quatCompR{w} - \quatCompK{a} \quatCompI{w} ) \imagJ
    - (\quatCompR{a} \quatCompK{w} - \quatCompI{a} \quatCompJ{w} + \quatCompJ{a} \quatCompI{w} + \quatCompK{a} \quatCompR{w} ) \imagK \\
    &= \frac{1}{4} 
    [\quatCompR{a} - \quatCompI{a} \imagI + \quatCompJ{a} \imagJ - \quatCompK{a} \imagK
    + (-\quatCompR{a}\imagI - \quatCompI{a} - \quatCompJ{a}\imagK - \quatCompK{a}\imagJ) \imagI \\
    &+ (\quatCompR{a}\imagJ + \quatCompI{a}\imagK - \quatCompJ{a} - \quatCompK{a}\imagI) \imagJ
    + (-\quatCompR{a}\imagK + \quatCompI{a}\imagJ + \quatCompJ{a}\imagI - \quatCompK{a}) \imagK ] \\
    &= \frac{1}{2} \left[\quaternionConjComponents{a} \right] \\
    &= \frac{1}{2} \quatConj{a}
\end{aligned}    
\end{equation}
%
%
\begin{equation}
\begin{aligned}
    \frac{\partial \quatInvK{y}}{\partial \quatConj{w}} 
    &= \frac{\partial \InvK{(\quaternion{w}\quaternion{a} + \quaternion{b})}}{\partial \quatConj{w}}
    = \frac{\partial \InvK{(\quaternion{w}\quaternion{a})}}{\partial \quatConj{w}} \\
    &= \frac{\partial}{\partial \quatConj{w}} 
    (\quatCompR{a} \quatCompR{w} - \quatCompI{a} \quatCompI{w} - \quatCompJ{a} \quatCompJ{w} - \quatCompK{a} \quatCompK{w})
    - (\quatCompR{a} \quatCompI{w} + \quatCompI{a} \quatCompR{w} - \quatCompJ{a} \quatCompK{w} + \quatCompK{a} \quatCompJ{w} ) \imagI \\
    &- (\quatCompR{a} \quatCompJ{w} + \quatCompI{a} \quatCompK{w} + \quatCompJ{a} \quatCompR{w} - \quatCompK{a} \quatCompI{w} ) \imagJ
    + (\quatCompR{a} \quatCompK{w} - \quatCompI{a} \quatCompJ{w} + \quatCompJ{a} \quatCompI{w} + \quatCompK{a} \quatCompR{w} ) \imagK \\
    &= \frac{1}{4} 
    [\quatCompR{a} - \quatCompI{a} \imagI - \quatCompJ{a} \imagJ + \quatCompK{a} \imagK
    + (-\quatCompR{a}\imagI - \quatCompI{a} + \quatCompJ{a}\imagK + \quatCompK{a}\imagJ) \imagI \\
    &+ (-\quatCompR{a}\imagJ - \quatCompI{a}\imagK - \quatCompJ{a} - \quatCompK{a}\imagI) \imagJ
    + (\quatCompR{a}\imagK - \quatCompI{a}\imagJ + \quatCompJ{a}\imagI - \quatCompK{a}) \imagK ] \\
    &= \frac{1}{2} \left[\quaternionConjComponents{a} \right] \\
    &= \frac{1}{2} \quatConj{a}
\end{aligned}    
\end{equation}

Combining both parts to form the overall derivative yields 

\begin{equation}
\begin{aligned}
    \frac{\partial \loss(\quaternion{y}(\quaternion{w}, \quaternion{b})}{\partial \quatConj{w}} 
    &= \frac{1}{2}\left(\quatCompR{e} - \quatCompI{e}\imagI - \quatCompJ{e}\imagJ - \quatCompK{e}\imagK \right) \frac{1}{2} \quatConj{a} \\
    &+ \frac{1}{2}\left(\quatCompR{e} - \quatCompI{e}\imagI + \quatCompJ{e}\imagJ + \quatCompK{e}\imagK \right) (-\frac{1}{2} \quatConj{a}) \\
    &+ \frac{1}{2}\left(\quatCompR{e} + \quatCompI{e}\imagI - \quatCompJ{e}\imagJ + \quatCompK{e}\imagK \right) (-\frac{1}{2} \quatConj{a}) \\
    &+ \frac{1}{2}\left(\quatCompR{e} + \quatCompI{e}\imagI + \quatCompJ{e}\imagJ - \quatCompK{e}\imagK \right) (-\frac{1}{2} \quatConj{a}) \\
    &= \frac{1}{2} (\quatCompR{e} - \quatCompI{e}\imagI - \quatCompJ{e}\imagJ - \quatCompK{e}\imagK \\
    &~~~~~         -\quatCompR{e} + \quatCompI{e}\imagI - \quatCompJ{e}\imagJ - \quatCompK{e}\imagK \\
    &~~~~~         -\quatCompR{e} - \quatCompI{e}\imagI + \quatCompJ{e}\imagJ - \quatCompK{e}\imagK \\
    &~~~~~         -\quatCompR{e} - \quatCompI{e}\imagI - \quatCompJ{e}\imagJ + \quatCompK{e}\imagK )\frac{1}{2} \quatConj{a} \\
    &=\frac{1}{2} (-\quatCompR{e} -2\quatCompI{e}\imagI -2\quatCompJ{e}\imagJ -\quatCompK{e}\imagK) \frac{1}{2} \quatConj{a} \\
    &= -\frac{1}{2} \quaternion{e} \quatConj{a} = -\frac{1}{2} (\quaternion{d}-\quaternion{y}) \quatConj{a}
\end{aligned}
\end{equation}

Alternatively, this can also be calculated as 
\begin{equation}
\begin{aligned}
    \frac{\partial \loss(\quaternion{y}(\quaternion{w}, \quaternion{b})}{\partial \quatConj{w}} 
    &= \frac{-1}{2}\quatConj{e}\frac{-1}{2}\quatConj{a} + \frac{-1}{2}\quatConjInvI{e}\frac{1}{2}\quatConj{a} + \frac{-1}{2}\quatConjInvJ{e}\frac{1}{2}\quatConj{a}  + \frac{-1}{2}\quatConjInvK{e}\frac{1}{2}\quatConj{a} \\
    &= \frac{1}{4} \left[ \quatConj{e} - \quatConjInvI{e} - \quatConjInvJ{e} - \quatConjInvK{e} \right] \quatConj{a} \\
    &= -\frac{1}{2} \frac{1}{2} \left[-\quatConj{e} + \quatConjInvI{e} + \quatConjInvJ{e} + \quatConjInvK{e} \right] \quatConj{a} \\
    &= -\frac{1}{2} \quaternion{e}\quatConj{a}~~\text{(using \eqref{equ:quaternionInvolution5})}
\end{aligned}
\label{equ:last_layer_error_wrespect_w}
\end{equation}

which avoids the tedious sorting of the respective terms and highlights the convenience obtained by using Equations \eqref{equ:quaternionInvolution2}-\eqref{equ:quaternionInvolution5}.\\
\\

\subsubsection{Derivative with respect to bias}

Likewise, we can derive with respect to the bias $\quaternion{b}_i^{(L)}$

\begin{equation}
    \frac{\partial \loss(\quaternion{y}(\quaternion{w}, \quaternion{b})}{\partial \quatConj{b}} = 
    \quaternionDerivative{\loss}{\quaternion{y}}{\quatConj{b}}.
\end{equation}

The left partial derivatives are already known, hence we just need to calculate the right ones:

\begin{equation}
\begin{aligned}
    \frac{\partial \quaternion{y}}{\partial \quatConj{b}} 
    &= \frac{\partial (\quaternion{w}\quaternion{a} + \quaternion{b})}{\partial \quatConj{b}}
    = \frac{\partial (\quaternion{b})}{\partial \quatConj{b}} \\
    &= \frac{\partial}{\partial \quatConj{b}} \quatCompR{b} + \quatCompI{b} \imagI + \quatCompJ{b} \imagJ + \quatCompK{b} \imagK\\
    &= \frac{1}{4} \left[ 1 + \imagI\imagI + \imagJ\imagJ + \imagK\imagK \right] \\
    &= \frac{1}{4} \left[ 1 -1 -1 -1 \right] \\
    &= - 0.5
\end{aligned}    
\end{equation}

\begin{equation}
\begin{aligned}
    \frac{\partial \quatInvI{y}}{\partial \quatConj{b}} 
    &= \frac{\partial \InvI{(\quaternion{w}\quaternion{a} + \quaternion{b})}}{\partial \quatConj{b}}
    =  \frac{\partial \InvI{(\quaternion{b})}}{\partial \quatConj{b}} \\
    &= \frac{\partial}{\partial \quatConj{b}} \quatCompR{b} + \quatCompI{b} \imagI - \quatCompJ{b} \imagJ - \quatCompK{b} \imagK\\
    &= \frac{1}{4} \left[ 1 + \imagI\imagI - \imagJ\imagJ - \imagK\imagK \right] \\
    &= \frac{1}{4} \left[ 1 -1 +1 +1 \right] \\
    &= 0.5
\end{aligned}    
\end{equation}

\begin{equation}
\begin{aligned}
    \frac{\partial \quatInvJ{y}}{\partial \quatConj{b}} 
    &= \frac{\partial \InvJ{(\quaternion{w}\quaternion{a} + \quaternion{b})}}{\partial \quatConj{b}}
    =  \frac{\partial \InvJ{(\quaternion{b})}}{\partial \quatConj{b}} \\
    &= \frac{\partial}{\partial \quatConj{b}} \quatCompR{b} - \quatCompI{b} \imagI + \quatCompJ{b} \imagJ - \quatCompK{b} \imagK\\
    &= \frac{1}{4} \left[ 1 - \imagI\imagI + \imagJ\imagJ - \imagK\imagK \right] \\
    &= \frac{1}{4} \left[ 1 +1 -1 +1 \right] \\
    &= 0.5
\end{aligned}    
\end{equation}

\begin{equation}
\begin{aligned}
    \frac{\partial \quatInvK{y}}{\partial \quatConj{b}} 
    &= \frac{\partial \InvK{(\quaternion{w}\quaternion{a} + \quaternion{b})}}{\partial \quatConj{b}}
    =  \frac{\partial \InvK{(\quaternion{b})}}{\partial \quatConj{b}} \\
    &= \frac{\partial}{\partial \quatConj{b}} \quatCompR{b} - \quatCompI{b} \imagI - \quatCompJ{b} \imagJ + \quatCompK{b} \imagK\\
    &= \frac{1}{4} \left[ 1 - \imagI\imagI - \imagJ\imagJ + \imagK\imagK \right] \\
    &= \frac{1}{4} \left[ 1 +1 +1 -1 \right] \\
    &= 0.5
\end{aligned}    
\end{equation}

Combining and summing up the left and right partial derivatives is exactly the same as with the weights, but with a missing $a^*$ in the right terms. Consequently, the final derivative is

\begin{equation}
\begin{aligned}
    \frac{\partial \loss(\quaternion{e}(\quaternion{w}, \quaternion{b})}{\partial \quatConj{b}} = -\frac{1}{2} \quaternion{e} .
\end{aligned}
\end{equation}

\subsubsection{Derivative with respect to activations}
Finally, we need to derive the loss with respect to the activations / outputs of the previous layer $\quaternion{a}_j^{(L-1)}$. As multiple output neurons $i \in K$ are connected to $\quaternion{a}_j^{(L-1)}$, we have to take the sum of the respective derivatives:


\begin{equation}
\begin{aligned}
    \frac{\partial \loss(\quaternion{a}_j)}{\partial \quaternion{a}_j} 
    &= \sum_{i \in K} \frac{\partial \loss(\quaternion{y}_k(\quaternion{a_j})}{\partial \quaternion{a}_j} \\
    &= \sum_{i \in K} \quaternionDerivative{\loss}{\quaternion{y}_k}{\quaternion{a_j}}
\end{aligned}
\end{equation}

The calculations for the individual parts of the sum are analog to the ones when deriving with respect to the weights, hence we will not list the detailed calculations here.

\begin{equation}
\begin{aligned}
    \frac{\partial \quaternion{y}}{\partial \quaternion{a}} 
    &= \frac{\partial (\quaternion{w}\quaternion{a} + \quaternion{b})}{\partial \quaternion{a}}  \\
    &= \quaternion{w}
\end{aligned}    
\label{equ:right_partial_derivative_a_r}
\end{equation}

\begin{equation}
\begin{aligned}
    \frac{\partial \quatInvI{y}}{\partial \quaternion{a}} 
    &= \frac{\partial \InvI{(\quaternion{w}\quaternion{a} + \quaternion{b})}}{\partial \quaternion{a}}  \\
    &= 0
\end{aligned}    
\end{equation}

\begin{equation}
\begin{aligned}
    \frac{\partial \quatInvJ{y}}{\partial \quaternion{a}} 
    &= \frac{\partial \InvJ{(\quaternion{w}\quaternion{a} + \quaternion{b})}}{\partial \quaternion{a}}  \\
    &= 0
\end{aligned}    
\end{equation}

\begin{equation}
\begin{aligned}
    \frac{\partial \quatInvK{y}}{\partial\quaternion{ }a}
    &= \frac{\partial \InvK{(\quaternion{w}\quaternion{a} + \quaternion{b})}}{\partial \quaternion{a}}  \\
    &= 0
\end{aligned}    
\label{equ:right_partial_derivative_a_k}
\end{equation}

Now the final derivative formulates as


\begin{equation}
    \frac{\partial \loss(\quaternion{a}_j)}{\partial \quaternion{a}_j} 
    = \sum_{i \in K} -\frac{1}{2}\quatConj{e}_i \quaternion{w}_{i, j} .
\end{equation}

\subsubsection{Update rules for the last layer}

Based on the previous calculations, the update rules of weights and biases at timestep $n$ for the last layer are

%
\begin{equation}
\begin{aligned}
   \quaternion{w}_{i, j}^{(L)}(n + 1) &= \quaternion{w}_{i, j}^{(L)}(n) + \lambda \frac{1}{2} \quaternion{e}_i(n){\quatConj{a}}_j^{(L-1)}(n) \\
   \quaternion{b}_i^{(L)}(n + 1) &= \quaternion{b}_i^{(L)}(n) + \lambda \frac{1}{2} \quaternion{e}_i(n) .
\end{aligned}
\end{equation}

\subsection{Hidden layer}
\label{subsec:hidden_layer}

For the hidden layers, a litte more work needs to be done, especially as we now usually have an activation function and hence three parts we can derive for. As we know already from the quaternion chain rule, we cannot simply multiplicatively combine them, especially for the three components. Instead, we first start with deriving with respect the activation input where the loss is a function $\loss(\quaternion{a}^{(l)}) = \loss\left(\quaternion{a}^{(l)} (\quaternion{z}^{(l)}) \right); \quaternion{a}^{(l)}=\sigma (\quaternion{z}^{(l)})$. 

Then, we can create the involutions of this derivative and continue with deriving with respect to $\quaternion{w}_{i, j}^{(l)}$, $\quaternion{b}_{i}^{(l)}$ and $\quaternion{a}_{i}^{(l-1)}$. For simplicity and better readability, we will avoid the superscript $\square^{(l)}$ indicating the layer when we deal with the involutions to prevent double superscripts.

\subsubsection{Derivative with respect to the activation input}

For deriving with respect to the activation input $\quaternion{z}_i^{(l)}$ we need to calculate

\begin{equation}
\begin{aligned}
    \frac{\partial \loss}{\partial \quaternion{z}^{(l)}} 
    &= \frac{\partial \loss\left(\quaternion{a}^{(l)} (\quaternion{z}^{(l)})\right)}{\partial \quaternion{z}^{(l)}} \\
    &= \quaternionDerivative{\loss}{\quaternion{a}}{\quaternion{z}}.
\end{aligned}
\end{equation}

As indicated by Equations \eqref{equ:quaternion_chain_rule_1} and \eqref{equ:quaternion_chain_rule_2}, we can derive the outer equation with respect to both, the regular quaternion or it's conjugate, and here it's more convenient to chose the regular quaternion as $\quaternion{z}$ is a direct result of the forward phase and $\quatConj{z}$ is not.


To get the derivatives with respect to the involutions $\quatInvI{a},~\quatInvJ{a}~\text{and}~\quatInvK{a}$ we can simply take the known result and flip the signs of the imaginary parts according to Equation \eqref{equ:hr_calculus}.
For the last layer we know $\frac{\partial \loss}{\partial \quaternion{a}}$ already. For an arbitrary hidden layer $(l)$ we don't know it yet. Hence, for better readability and generalization, in the following we will call the result simply $ \quaternion{p}^{(l+1)} = \quaternionConjComponents{p}$. As this is coming from the following layer we assign the superscript $\square^{(l+1)}$ Furthermore, this naming is also convenient since the result will change from the last layer to the hidden layer, but we can always refer to it as $\quaternion{p}$. Hence 

\begin{equation}
\begin{aligned}
    \frac{\partial \quaternion{y}}{\partial \quatInvI{a}} = \quatCompR{p} + \quatCompI{p} \imagI - \quatCompJ{p} \imagJ - \quatCompK{p} \imagK \\
    \frac{\partial \quaternion{y}}{\partial \quatInvJ{a}} = \quatCompR{p} - \quatCompI{p} \imagI + \quatCompJ{p} \imagJ - \quatCompK{p} \imagK \\
    \frac{\partial \quaternion{y}}{\partial \quatInvK{a}} = \quatCompR{p} - \quatCompI{p} \imagI - \quatCompJ{p} \imagJ + \quatCompK{p} \imagK.\\
\end{aligned}
\end{equation}

For the right part of the partial derivatives, we first need to know $a(z)$ and it's involutions $a^{i}(z),~a^{j}(z)\text{and}~a^{k}(z)$. Using an element-wise operating activation function $\sigma(\cdot)$ to calculate $a^{(l)} = \sigma(z) = \sigma(z_0) + \sigma(z_1) i + \sigma(z_2) j + \sigma(z_3) k$ these are

\begin{equation}
\begin{aligned}
    \quaternion{a} = \sigma(\quaternion{z}) &= \sigma(\quatCompR{z}) + \sigma(\quatCompI{z})\imagI + \sigma(\quatCompJ{z})\imagJ + \sigma(\quatCompK{z})\imagK \\
    \quatInvI{a}   = \sigma(\quaternion{z}) &= \sigma(\quatCompR{z}) + \sigma(\quatCompI{z})\imagI - \sigma(\quatCompJ{z})\imagJ - \sigma(\quatCompK{z})\imagK \\
    \quatInvJ{a}   = \sigma(\quaternion{z}) &= \sigma(\quatCompR{z}) - \sigma(\quatCompI{z})\imagI + \sigma(\quatCompJ{z})\imagJ - \sigma(\quatCompK{z})\imagK \\
    \quatInvK{a}   = \sigma(\quaternion{z}) &= \sigma(\quatCompR{z}) - \sigma(\quatCompI{z})\imagI - \sigma(\quatCompJ{z})\imagJ + \sigma(\quatCompK{z})\imagK .\\
\end{aligned}
\end{equation}

Hence, we get the derivations



\begin{equation}
\begin{aligned}
    \frac{\partial \quaternion{a}}{\partial \quaternion{z}} &= \frac{\partial}{\partial \quaternion{z}}
    \sigma(\quatCompR{z}) + \sigma(\quatCompI{z}) \imagI + \sigma(\quatCompJ{z}) \imagJ + \sigma(\quatCompK{z}) \imagK \\
    &=\frac{1}{4} \left[\sigma^{\prime}(\quatCompR{z}) - \sigma^{\prime}(\quatCompI{z}) \imagI\imagI - \sigma^{\prime}(\quatCompJ{z}) \imagJ\imagJ - \sigma^{\prime}(\quatCompK{z}) \imagK\imagK \right]\\
    &=\frac{1}{4} \left[\sigma^{\prime}(\quatCompR{z}) + \sigma^{\prime}(\quatCompI{z}) + \sigma^{\prime}(\quatCompJ{z}) + \sigma^{\prime}(\quatCompK{z}) \right]
\end{aligned}
\end{equation}
\begin{equation}
\begin{aligned}
    \frac{\partial \quatInvI{a}}{\partial \quaternion{z}} &= \frac{\partial}{\partial \quaternion{z}}
    \sigma(\quatCompR{z}) + \sigma(\quatCompI{z}) \imagI - \sigma(\quatCompJ{z}) \imagJ - \sigma(\quatCompK{z}) \imagK \\
    &=\frac{1}{4} \left[\sigma^{\prime}(\quatCompR{z}) - \sigma^{\prime}(\quatCompI{z}) \imagI\imagI + \sigma^{\prime}(\quatCompJ{z}) \imagJ\imagJ + \sigma^{\prime}(\quatCompK{z}) \imagK\imagK \right]\\
    &=\frac{1}{4} \left[\sigma^{\prime}(\quatCompR{z}) + \sigma^{\prime}(\quatCompI{z}) - \sigma^{\prime}(\quatCompJ{z}) - \sigma^{\prime}(\quatCompK{z}) \right]
\end{aligned}
\end{equation}
\begin{equation}
\begin{aligned}
    \frac{\partial \quatInvJ{a}}{\partial \quaternion{z}} &= \frac{\partial}{\partial \quaternion{z}}
    \sigma(\quatCompR{z}) - \sigma(\quatCompI{z}) \imagI + \sigma(\quatCompJ{z}) \imagJ - \sigma(\quatCompK{z}) \imagK \\
    &=\frac{1}{4} \left[\sigma^{\prime}(\quatCompR{z}) + \sigma^{\prime}(\quatCompI{z}) \imagI\imagI - \sigma^{\prime}(\quatCompJ{z}) \imagJ\imagJ + \sigma^{\prime}(\quatCompK{z}) \imagK\imagK \right]\\
    &=\frac{1}{4} \left[\sigma^{\prime}(\quatCompR{z}) - \sigma^{\prime}(\quatCompI{z}) + \sigma^{\prime}(\quatCompJ{z}) - \sigma^{\prime}(\quatCompK{z}) \right]
\end{aligned}
\end{equation}
\begin{equation}
\begin{aligned}
    \frac{\partial \quatInvK{a}}{\partial \quaternion{z}} &= \frac{\partial}{\partial \quaternion{z}}
    \sigma(\quatCompR{z}) - \sigma(\quatCompI{z}) \imagI - \sigma(\quatCompJ{z}) \imagJ + \sigma(\quatCompK{z}) \imagK \\
    &=\frac{1}{4} \left[\sigma^{\prime}(\quatCompR{z}) + \sigma^{\prime}(\quatCompI{z}) \imagI\imagI + \sigma^{\prime}(\quatCompJ{z}) \imagJ\imagJ - \sigma^{\prime}(\quatCompK{z}) \imagK\imagK \right]\\
    &=\frac{1}{4} \left[\sigma^{\prime}(\quatCompR{z}) - \sigma^{\prime}(\quatCompI{z}) - \sigma^{\prime}(\quatCompJ{z}) + \sigma^{\prime}(\quatCompK{z}) \right]
\end{aligned}
\end{equation}

Combining the respective partial derivatives yields



\begin{equation}
\begin{aligned}
    \frac{\partial \loss}{\partial z^{(l)}} 
    &= \frac{\partial \loss\left(a^{(l)} (z^{(l)})\right)}{\partial z^{(l)}} \\
    &= \quaternion{p} \frac{1}{4} \left[\sigma^{\prime}(\quatCompR{z}) + \sigma^{\prime}(\quatCompI{z}) + \sigma^{\prime}(\quatCompJ{z}) + \sigma^{\prime}(\quatCompK{z}) \right] \\
    &+ \quatInvI{p} \frac{1}{4} \left[\sigma^{\prime}(\quatCompR{z}) + \sigma^{\prime}(\quatCompI{z}) - \sigma^{\prime}(\quatCompJ{z}) - \sigma^{\prime}(\quatCompK{z}) \right] \\
    &+ \quatInvJ{p} \frac{1}{4} \left[\sigma^{\prime}(\quatCompR{z}) - \sigma^{\prime}(\quatCompI{z}) + \sigma^{\prime}(\quatCompJ{z}) - \sigma^{\prime}(\quatCompK{z}) \right] \\
    &+ \quatInvK{p} \frac{1}{4} \left[\sigma^{\prime}(\quatCompR{z}) - \sigma^{\prime}(\quatCompI{z}) - \sigma^{\prime}(\quatCompJ{z}) + \sigma^{\prime}(\quatCompK{z}) \right] \\
    &= \frac{1}{4} \left[\quaternion{p} + \quatInvI{p} + \quatInvJ{p} +\quatInvK{p} \right] \sigma^{\prime}(\quatCompR{z}) \\
    &+ \frac{1}{4} \left[\quaternion{p} + \quatInvI{p} - \quatInvJ{p} -\quatInvK{p} \right] \sigma^{\prime}(\quatCompI{z}) \\
    &+ \frac{1}{4} \left[\quaternion{p} - \quatInvI{p} + \quatInvJ{p} -\quatInvK{p} \right] \sigma^{\prime}(\quatCompJ{z}) \\
    &+ \frac{1}{4} \left[\quaternion{p} - \quatInvI{p} - \quatInvJ{p} +\quatInvK{p} \right] \sigma^{\prime}(\quatCompK{z}) \\
    &= p_0 \sigma^{\prime}(z_0) + p_1 \sigma^{\prime}(z_1) i + p_2 \sigma^{\prime}(z_2) j + p_3 \sigma^{\prime}(z_3) k \\
    &= \quaternion{p} \circ \sigma^{\prime}(\quaternion{z})
\end{aligned}
\end{equation}

Note that in the last step we used the alternative representations
\begin{equation}
\begin{aligned}
    \quatCompR{q}        &= \frac{1}{4} \left( \quaternion{q} + \quatInvI{q} + \quatInvJ{q} + \quatInvK{q} \right)
    &\quatCompI{q} \imagI = \frac{1}{4} \left( \quaternion{q} + \quatInvI{q} - \quatInvJ{q} - \quatInvK{q} \right) \\
    \quatCompJ{q} \imagJ &= \frac{1}{4} \left( \quaternion{q} - \quatInvI{q} + \quatInvJ{q} - \quatInvK{q} \right)
    &\quatCompK{q} \imagK = \frac{1}{4} \left( \quaternion{q} - \quatInvI{q} - \quatInvJ{q} + \quatInvK{q} \right) \\
\end{aligned}
\end{equation}
of \eqref{equ:quaternionInvolution2} which can be obtained by multiplying with $\{1, \imagI, \imagJ, \imagK\}$ respectively.

\subsubsection{Derivative with respect to the weights}

Now we can continue with deriving

\begin{equation}
\begin{aligned}
    \frac{\partial \loss(w)}{\partial \quaternion{w}^{*^{(l)}}} &= \frac{\partial \loss(\quaternion{z}^{(l)}(\quaternion{w}^{(l)}))}{\partial \quaternion{w}^{*^{(l)}}} \\
    &= \frac{\partial \loss}{\partial z} \frac{\partial z}{\partial w^*}
    + \frac{\partial \loss}{\partial z^i} \frac{\partial z^i}{\partial w^*}
    + \frac{\partial \loss}{\partial z^j} \frac{\partial z^j}{\partial w^*}
    + \frac{\partial \loss}{\partial z^k} \frac{\partial z^k}{\partial w^*}
\end{aligned}
\end{equation}

Just like in the case above we assign a name $\frac{\partial \loss}{\partial \quaternion{z}^{(l)}} = \quaternion{q} = \quaternionConjComponents{q}$ for the result. 
Then, we get the following partial derivatives for the involutions:


\begin{equation}
\begin{aligned}
    \frac{\partial \quaternion{y}}{\partial \quatInvI{z}} = \quatCompR{q} + \quatCompI{q} \imagI - \quatCompJ{q} \imagJ - \quatCompK{q} \imagK = \quatInvI{q} \\
    \frac{\partial \quaternion{y}}{\partial \quatInvJ{z}} = \quatCompR{q} - \quatCompI{q} \imagI + \quatCompJ{q} \imagJ - \quatCompK{q} \imagK = \quatInvJ{q} \\
    \frac{\partial \quaternion{y}}{\partial \quatInvK{z}} = \quatCompR{q} - \quatCompI{q} \imagI - \quatCompJ{q} \imagJ + \quatCompK{q} \imagK = \quatInvK{q} \\
\end{aligned}
\end{equation}

The right partial derivatives with respect to the weights are already known from Subsection \ref{subsec:final_layer}, namely 
\begin{equation}
\begin{aligned}
    \frac{\partial \quaternion{z}}{\partial \quatConj{w}} = -\frac{1}{2}\quatConj{a},~~
    \frac{\partial \quatInvI{z}}{\partial \quatConj{w}} =  \frac{1}{2}\quatConj{a},~~
    \frac{\partial \quatInvJ{z}}{\partial \quatConj{w}} =  \frac{1}{2}\quatConj{a},~~
    \frac{\partial \quatInvK{z}}{\partial \quatConj{w}} =  \frac{1}{2}\quatConj{a}
\end{aligned}
\end{equation}
whereby $\quaternion{z}=\quaternion{z}_i^{(l)}$, $\quaternion{w}=\quaternion{w}_{i,j}^{(l)}$ and $\quaternion{a}=\quaternion{a}_i^{(l-1)}$. Consequently, the full derivation is 


\begin{equation}
\begin{aligned}
    \frac{\partial \loss(\quaternion{w})}{\partial \quatConj{w}} &= \frac{\partial \loss(\quaternion{z}(\quaternion{w}))}{\partial\quatConj{w}} \\
    &= \quaternion{q} \frac{-1}{2}\quatConj{a} + \quatInvI{q}\frac{1}{2}\quatConj{a} + \quatInvJ{q}\frac{1}{2}\quatConj{a} + \quatInvK{q}\frac{1}{2}\quatConj{a} \\
    &= \frac{1}{2} (-\quaternion{q} + \quatInvI{q} + \quatInvJ{q} + \quatInvK{q}) \quatConj{a} \\
    &= \quatConj{q} \quatConj{a} ~\text{(using \eqref{equ:quaternionInvolution4})} .
\end{aligned}
\end{equation}

Note the similarities to Equation \eqref{equ:last_layer_error_wrespect_w}. 

\subsubsection{Derivative with respect to the biases}

In the same manner, we can also compute the derivation with respect to the bias $\quaternion{b}^{(l)}$
with the known results
\begin{equation}
\begin{aligned}
    \frac{\partial \quaternion{z}}{\partial \quatConj{b}} = -\frac{1}{2},~~
    \frac{\partial \quatInvI{z}}{\partial \quatConj{b}} =  \frac{1}{2},~~
    \frac{\partial \quatInvJ{z}}{\partial \quatConj{b}} =  \frac{1}{2},~~
    \frac{\partial \quatInvK{z}}{\partial \quatConj{b}} =  \frac{1}{2}
\end{aligned}
\end{equation}
where again we can find similarities to the derivation for the last layer:


\begin{equation}
\begin{aligned}
    \frac{\partial \loss(\quaternion{b})}{\partial \quatConj{b}} &= \frac{\partial \loss(\quaternion{z}(\quaternion{b}))}{\partial \quatConj{b}} \\
    &= \quaternion{q} \frac{-1}{2} + \quatInvI{q}\frac{1}{2} + \quatInvJ{q}\frac{1}{2} + \quatInvK{q}\frac{1}{2} \\
    &= \frac{1}{2} (-\quaternion{q} + \quatInvI{q} + \quatInvJ{q} + \quatInvK{q}) \\
    &= \quatConj{q} ~\text{(using \eqref{equ:quaternionInvolution4})}.
\end{aligned}
\end{equation}

\subsubsection{Derivative with respect to the activations of the previous layer}

Finally, we need to derive with respect to the activation of the previous layer $\quaternion{a}_j^{(l-1)}$ where again the right partial derivatives are already known from Equations \eqref{equ:right_partial_derivative_a_r} - \eqref{equ:right_partial_derivative_a_k}.
As in the regular backpropagation, in this case we need to consider all neurons $K$ where a respective activation $\quaternion{a}_j^{(l-1)}$ is input to.
Hence, the final derivative formulates as

\begin{equation}
\begin{aligned}
    \frac{\partial \loss(\quaternion{a}_j^{(l-1)})}{\partial \quaternion{a}_j^{(l-1)}} 
    &= \sum_{i \in K}
    \frac{\partial \loss(\quaternion{z}_i^{(l)}(\quaternion{a}_j^{(l-1)}))}{\partial \quaternion{a}_j^{(l-1)}} \\
    &= \sum_{i \in K} \quaternionDerivative{\loss}{\quaternion{z}_i}{\quaternion{a}_j} \\
    &= \sum_{i \in K} \quaternion{q}_i * \quaternion{w}_{i, j} + \quatInvI{q}_i * 0 + \quatInvJ{q}_i * 0 +\quatInvK{q}_i * 0 \\
    &= \sum_{i \in K} \quaternion{q}_i \quaternion{w}_{i, j}
\end{aligned}
\end{equation}


Note that the superscripts $\square^{\imagI, \imagJ, \imagK}$ indicate the quaternion involutions and the subscripts $\square_{i, j}$ the indices within the neural network. This achieved result will become the new $\quaternion{p}$ as a starting point for the derivatives of the previous layer.

\subsubsection{Update rules for the hidden layer}

Using the derived results, we can formulate the update rules for the parameters in the hidden layers as

\begin{equation}
\begin{aligned}
    \quaternion{w}_{i,j}^{(l)}(n + 1) &= \quaternion{w}_{i,j}^{(l)}(n) + \lambda \Bigl(
      {\quatCompR{p}}_i^{(l+1)} \sigma^{\prime}({\quatCompR{z}}_i^{(l)}) 
    + {\quatCompI{p}}_i^{(l+1)} \sigma^{\prime}({\quatCompI{z}}_i^{(l)}) \imagI \\
    &\qquad \qquad \qquad \qquad  + {\quatCompJ{p}}_i^{(l+1)} \sigma^{\prime}({\quatCompJ{z}}_i^{(l)}) \imagJ 
    + {\quatCompK{p}}_i^{(l+1)} \sigma^{\prime}({\quatCompK{z}}_i^{(l)}) \imagK
    \conj{\Bigr)} {\quatConj{a}}_j^{(l-1)} \\
    &= \quaternion{w}_{i,j}^{(l)}(n) + \lambda \left(
    \quaternion{p}_i^{(l)} \circ \sigma^{\prime}(\quaternion{z}_i^{(l)})  
    \right) {\quatConj{a}}_j^{(l-1)}\\
    \quaternion{b}_i^{(l)}(n + 1) &= \quaternion{b}_i^{(l)}(n) + \lambda \Bigl(
      {\quatCompR{p}}_i^{(l+1)} \sigma^{\prime}({\quatCompR{z}}_i^{(l)}) 
    + {\quatCompI{p}}_i^{(l+1)} \sigma^{\prime}({\quatCompI{z}}_i^{(l)}) \imagI \\
    & \qquad \qquad \qquad \qquad + {\quatCompJ{p}}_i^{(l+1)} \sigma^{\prime}({\quatCompJ{z}}_i^{(l)}) \imagJ 
    + {\quatCompK{p}}_i^{(l+1)} \sigma^{\prime}({\quatCompK{z}}_i^{(l)}) \imagK
    \conj{\Bigr)} \\
    &= \quaternion{b}_i^{(l)}(n) + \lambda \conj{\left(
    \quaternion{p}_i^{(l)} \circ \sigma^{\prime}(\quaternion{z}_i^{(l)})
    \right)} .
\end{aligned}
\end{equation}

Furthermore, the new $\quaternion{p}_j^{(l)}$ becomes


\begin{equation}
\begin{aligned}
    \quaternion{p}_j^{(l)} 
    &= \sum_{i \in K} \left( 
      {\quatCompR{p}}_i^{(l+1)} \sigma^{\prime}({\quatCompR{z}}_i^{(l)}) 
    + {\quatCompI{p}}_i^{(l+1)} \sigma^{\prime}({\quatCompI{z}}_i^{(l)}) \imagI
    + {\quatCompJ{p}}_i^{(l+1)} \sigma^{\prime}({\quatCompJ{z}}_i^{(l)}) \imagJ 
    + {\quatCompK{p}}_i^{(l+1)} \sigma^{\prime}({\quatCompK{z}}_i^{(l)}) \imagK
    \right)\quaternion{w}_{i, j}^{(l)} \\
    &= \sum_{i \in K} 
    \left(\quaternion{p}_i^{(l+1)} \circ \sigma^{\prime}(\quaternion{z}_i^{(l)})\right) \quaternion{w}_{i, j}^{(l)} .
\end{aligned}
\end{equation}
\section{Experiments}
\label{sec:experiments}

In this section, we will experimentally prove the convergence of the developed backpropagation algorithm. To do so, we create a synthetic quaternion valued dataset. This offers two advantages: First, in contrast to real world datasets we are ensured to have a perfect input to output relationship with a maximum error in the range of the numerical precision of the calculations used. Second, we have full control over the size of the dataset and it's numerical values to create a setup eliminating every possible disturbances potentially hurting the optimization procedure. Specifically, we use a quaternion valued model with three inputs, three layers with 3/2/2 outputs respectively and a Tanhshrink-activation in between them. The weights are randomly initialized such that $\lVert w_{i, j}^{(l)} \rVert = 1$ and we create 40000 random quaternion valued inputs for training and 10000 for validation.

\begin{figure}[h!]
    \centering
    \includegraphics[width=0.8\textwidth]{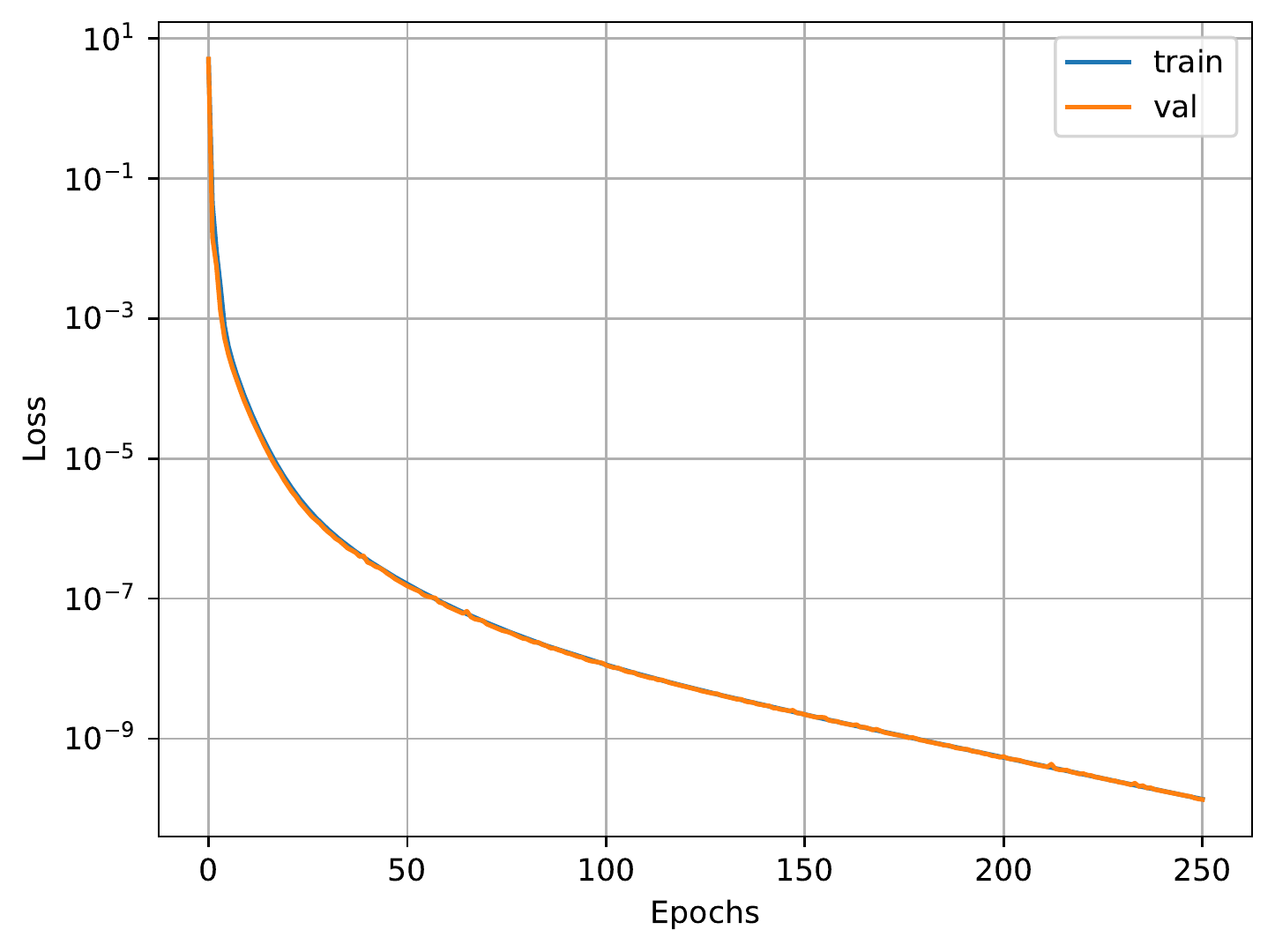}
    \caption{Loss-Graph from training the NN}
    \label{fig:loss}
\end{figure}

Then we train a second model with the same architecture, but different randomly initialized weights on this dataset. Here, we trained for 250 epochs using plain SGD, a learning rate of $1 \times 10^{-1}$ and a batch-size of 32. This yielded the loss trajectory as shown in Figure \ref{fig:loss}, with a final loss of $1.35 \times 10^{-10}$.

\begin{figure}[h!]
    \centering
    \includegraphics[width=0.95\textwidth]{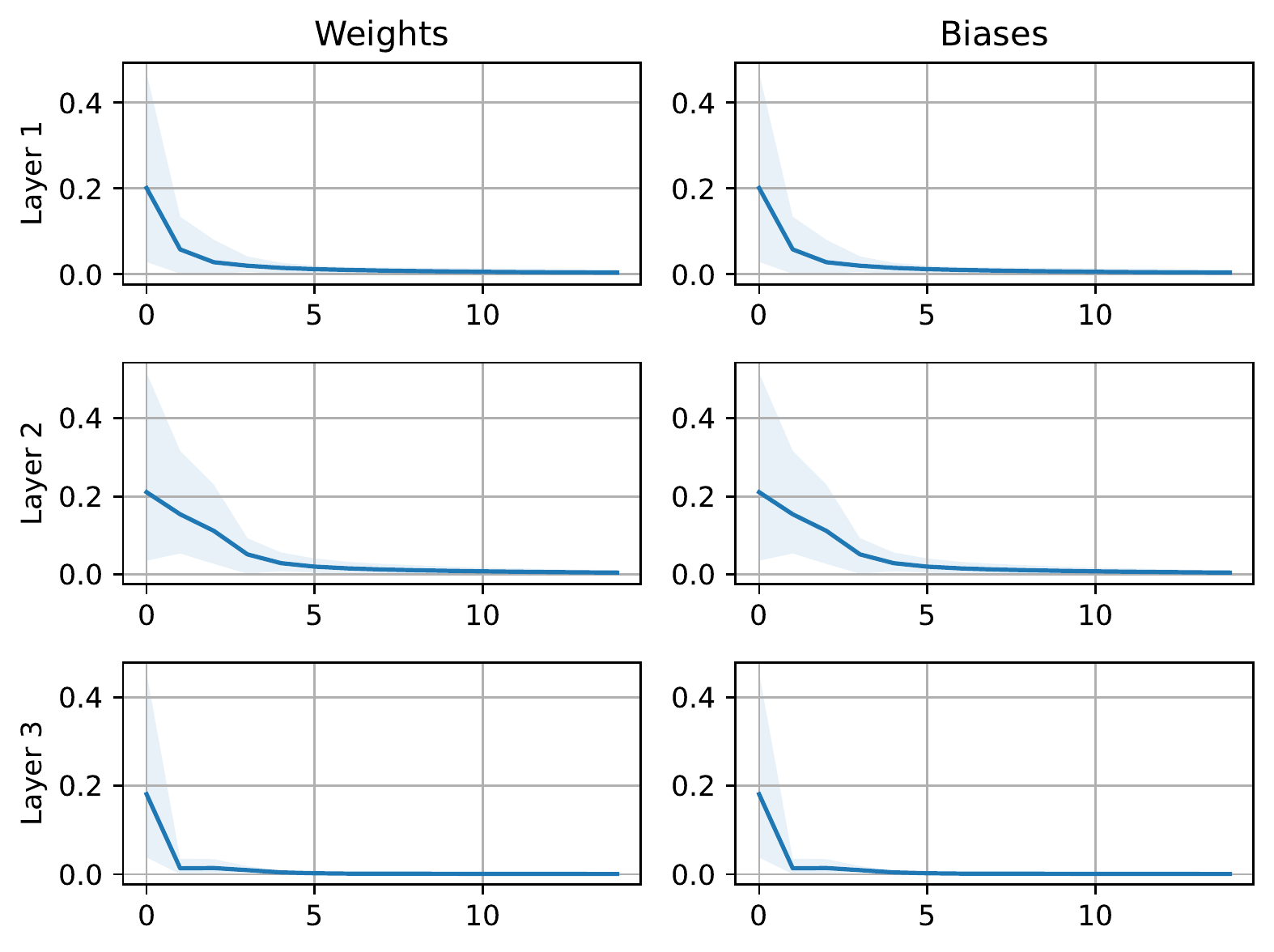}
    \caption{Difference of weights and biases to the ground truth model. The thick line corresponds to the mean difference to the ground truth weights, the light blue area indicates the range of minimum and maximum difference. We only show the first 15 epochs here as the difference is no longer perceptible afterwards.}
    \label{fig:weight_diffs}
\end{figure}

We observe that indeed the model trained using the developed quaternion backpropagation is capable of minimizing the error and converging to zero, proving the effectiveness. Furthermore, we can also compare the weights of the trained model with the ground-truth model as shown in Figure \ref{fig:weight_diffs}.

Again we can see a convergence to a difference of zero and hence to the ground-truth.

\section{Conclusion}
\label{sec:conclusion}

In this paper, we developed a novel quaternion backpropagation utilizing the GHR-Calculus. After introducing the required fundamentals and quaternion maths, we showed that by using plain partial derivatives with respect to the quaternion components as in other approaches to quaternion backpropagation, the product and more critical, the chain rule, does not hold. By applying the GHR calculus, we end up with derivatives which do, to create our quaternion backpropagation algorithm. This optimization is then successfully experimentally proven.

\printbibliography

\end{document}